\documentclass[twocolumn]{svjour3}
\smartqed
\usepackage{graphicx}
\usepackage{float}
\usepackage{times}
\usepackage{amsmath}
\usepackage{amsfonts}
\usepackage{amssymb}
\usepackage{multirow}
\usepackage{color}
\usepackage{tabu}
\usepackage[authoryear,sort&compress,]{natbib}
\usepackage[colorlinks=true,urlcolor=blue,citecolor=blue,linkcolor=blue,bookmarks=true]{hyperref}
\usepackage[title]{appendix}
\usepackage[english]{babel}
\usepackage{bm}
\usepackage{colortbl}
\usepackage{enumitem}
\usepackage{array}
\usepackage[ruled,vlined]{algorithm2e}
\usepackage{threeparttable}
\usepackage{dsfont}
\usepackage{pifont}
\usepackage{subcaption}
\usepackage{nccmath}
\usepackage{booktabs}
\usepackage{listings,lstautogobble}
\usepackage{multirow}
\usepackage{dsfont}
\usepackage{pifont}
\usepackage{marvosym}
\newcommand{\cmark}{\ding{51}}%
\newcommand{\xmark}{\ding{55}}%

\graphicspath{{./fig/intro/}{./fig/approach/}{./fig/exp/}{./}}

\newcommand{\bnInplaceSync}{\textsc{InPlace-ABN$^{\mathsf{sync}}$}\xspace}

\definecolor{codegreen}{rgb}{0,0.5,0}
\definecolor{codeblue}{rgb}{0.25,0.5,0.5}
\definecolor{codegray}{rgb}{0.6,0.6,0.6}

\makeatletter\renewcommand\paragraph{\@startsection{paragraph}{4}{\z@}
	{.5em \@plus1ex \@minus.2ex}{-.5em}{\normalfont\normalsize\bfseries}}\makeatother
\newlength\savewidth\newcommand\shline{\noalign{\global\savewidth\arrayrulewidth
	\global\arrayrulewidth 1.5pt}\hline\noalign{\global\arrayrulewidth\savewidth}}

\newcommand{\etal}{\textit{et al}.}

\journalname{IJCV}
\begin{document}\sloppy
	
	\title{OCNet: Object Context for Semantic Segmentation}
	
	\titlerunning{OCNet: Object Context for Semantic Segmentation}        %
	\author{
		Yuhui Yuan$^{\textbf{1},\textbf{3},\textbf{4}}$\and
		Lang Huang$^\textbf{2}$\and 
		Jianyuan Guo$^\textbf{2}$\and 
		Chao Zhang$^\textbf{2}$\and \\
		Xilin Chen$^{\textbf{3},\textbf{4}}$\and
		Jingdong Wang$^\textbf{1}$
	}
	
	\authorrunning{Yuan et al.} %
	
	\institute{
		$^{\textbf{1}}$Microsoft Research Asia \\
		$^{\textbf{2}}$Peking University \\
		$^{\textbf{3}}$Institute of Computing Technology, CAS \\
		$^{\textbf{4}}$University of Chinese Academy of Sciences \\
		\Letter \quad yuyua@microsoft.com, jingdw@microsoft.com
	}
	
	\date{Received: date / Accepted: date}
	
	\maketitle

	\vspace{-2mm}
	\begin{abstract}
		 In this paper, we address the semantic segmentation task
		 with a new context aggregation
		 scheme named \emph{object context},
		 which focuses on enhancing the role of object information.
		 Motivated by the fact that the category of each pixel 
		 is inherited from the object it belongs to,
		 we define the object context for each pixel as 
		 the set of pixels that
		 belong to the same category as the given pixel in the image.
		 We use a binary relation matrix to represent
		 the relationship between all pixels, where the value one
		 indicates the two selected pixels belong to
		 the same category and zero otherwise.

		 We propose to use a dense relation matrix 
		 to serve as a surrogate for the binary relation matrix.
		 The dense relation matrix is capable to emphasize the contribution of object information as the relation scores tend to be larger on the object pixels than the other pixels.
		 Considering that the dense relation matrix estimation requires quadratic computation overhead and memory consumption
		 w.r.t. the input size,
		 we propose an efficient interlaced sparse self-attention scheme
		 to model the dense relations between any two of all pixels 
		 via the combination of two sparse relation matrices.
		 
		 To capture richer context information,
		 we further combine our interlaced sparse self-attention scheme 
		 with the conventional multi-scale context schemes including
		 pyramid pooling~\citep{zhao2017pyramid} and atrous spatial pyramid pooling~\citep{chen2018deeplab}.
		 We empirically show the advantages of our approach with
		 competitive performances on five challenging benchmarks
         including:
		 Cityscapes, ADE20K, LIP,
		 PASCAL-Context and COCO-Stuff. \
	 \end{abstract}
	
	\section{Introduction}
	\label{sec:introduction}

	Semantic segmentation is a fundamental topic in computer vision 
	and is critical for {\color{black}{various scene understanding problems.
	It is typically formulated as a task of predicting the category of each pixel,
	i.e., the category of the object that the pixel belongs to.}}
	We are mainly interested in improving
	the pixel classification accuracy through explicitly 
	identifying the object region that the pixel
	belongs to.
	
	Extensive efforts based on deep convolutional neural networks 
	have been made to address the semantic segmentation
	since the pioneering approach of the fully convolutional network (FCN)~\citep{long2015fully}.
	{\color{black}{
	The original FCN approach suffers from two main drawbacks including the 
	\emph{reduced feature resolution} that loses the detailed spatial information
	and the \emph{small effective receptive field} that fails to capture long-range dependencies.
	There exist two main paths to tackle the above drawbacks:
	}}
	(i) raising the resolution
	of feature maps for improving the spatial precision
	or maintaining a high-resolution response map {\color{black}{through all stages}},
	e.g., through dilated convolutions~\citep{chen2018deeplab, yu2015multi},
	decoder network~\citep{badrinarayanan2017segnet,ronneberger2015u}
	or high-resolution networks~\citep{SunXLW19,SunZJCXLMWLW19}. 
	(ii) exploiting the global context
	to capture long-range dependencies,
	e.g., ParseNet~\citep{liu2015parsenet}, DeepLabv3~\citep{chen2018deeplab},
	and PSPNet~\citep{zhao2017pyramid}.
	In this work,
	we focus on the second path
	and propose a more efficient context scheme.{\color{black}{
	We define the context of a pixel 
	as a set of selected pixels
	and its context representation as an aggregation of 
	all selected pixels'
	representations if not specified.
	}}

	Most previous representative studies
	mainly exploit the multi-scale context
	formed from spatially nearby or sampled pixels.
	For instance, 
	the pyramid pooling module (PPM) in PSPNet~\citep{zhao2017pyramid} 
	divides all pixels into multiple regions
	and selects all pixels lying in the same region with a pixel as its context.
	The atrous spatial pyramid pooling module (ASPP) in 
	DeepLabv3~\citep{chen2017rethinking}
	selects the surrounding pixels of a pixel
	with different dilation rates as its context.{\color{black}{
	Therefore, the selected pixels of both PPM context and ASPP context
	tend to be the mixture of object pixels, relevant background pixels and irrelevant background pixels.
	Motivated by the fact that category of each pixel is essentially the category of the object that it belongs to, we should enhance the object pixels that constitute the object.}}
	
	\begin{figure}
		\centering
		\includegraphics[width=.15\textwidth]{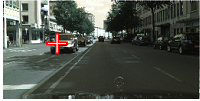}
		\includegraphics[width=.15\textwidth]{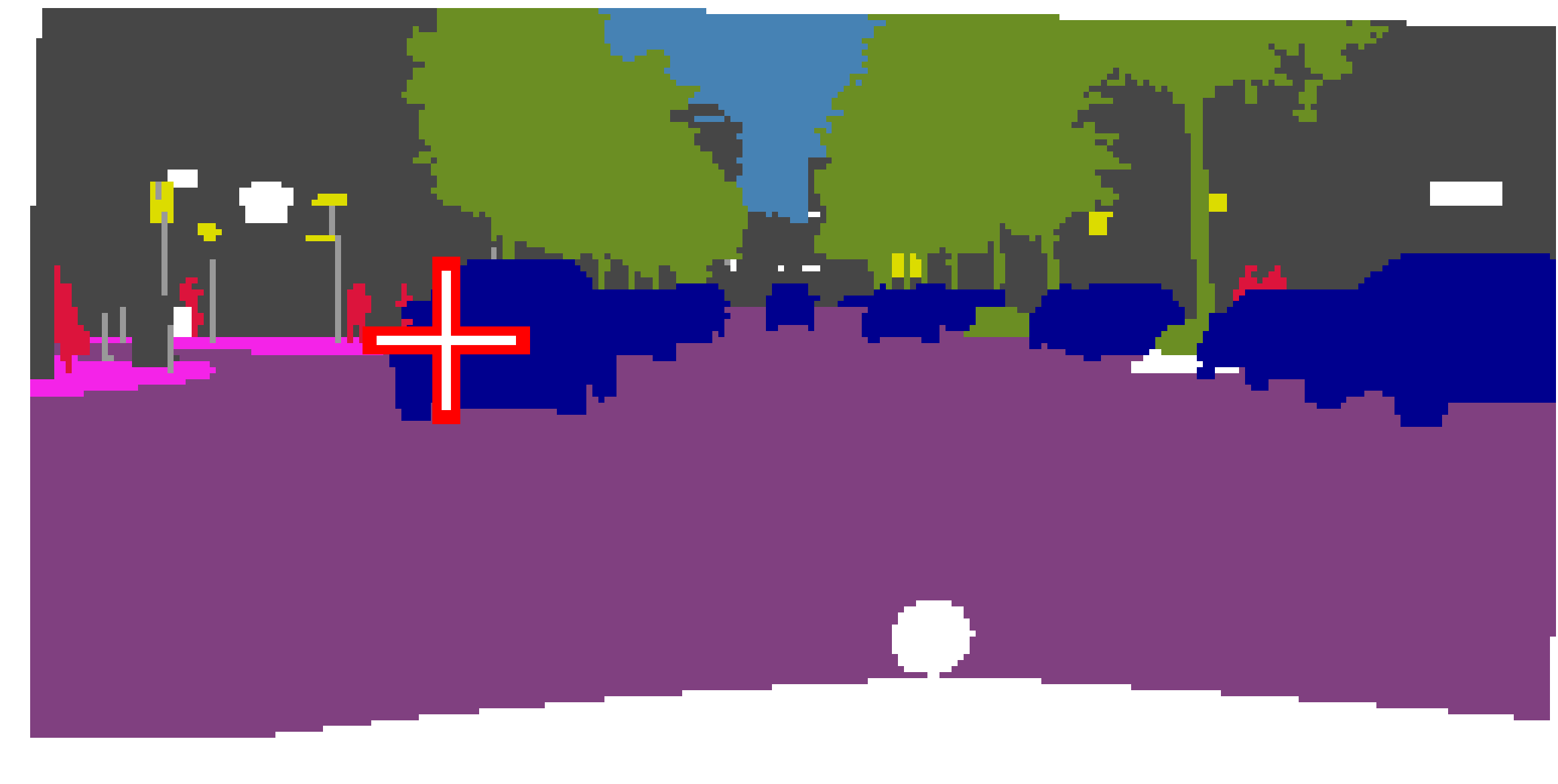}
		\includegraphics[width=.15\textwidth]{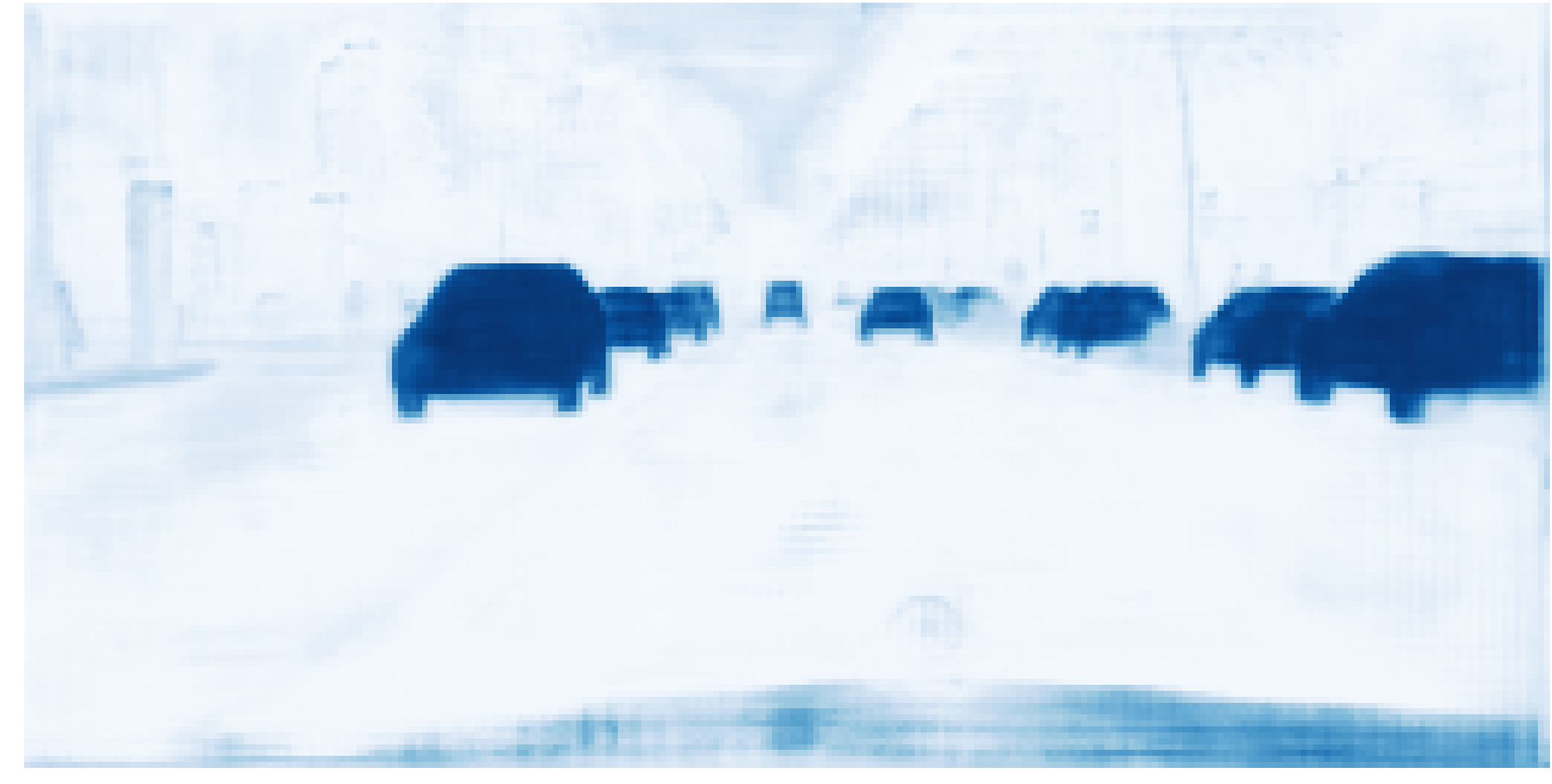}\\
		\vspace{.05cm}
		\includegraphics[width=.15\textwidth]{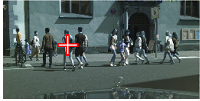}
		\includegraphics[width=.15\textwidth]{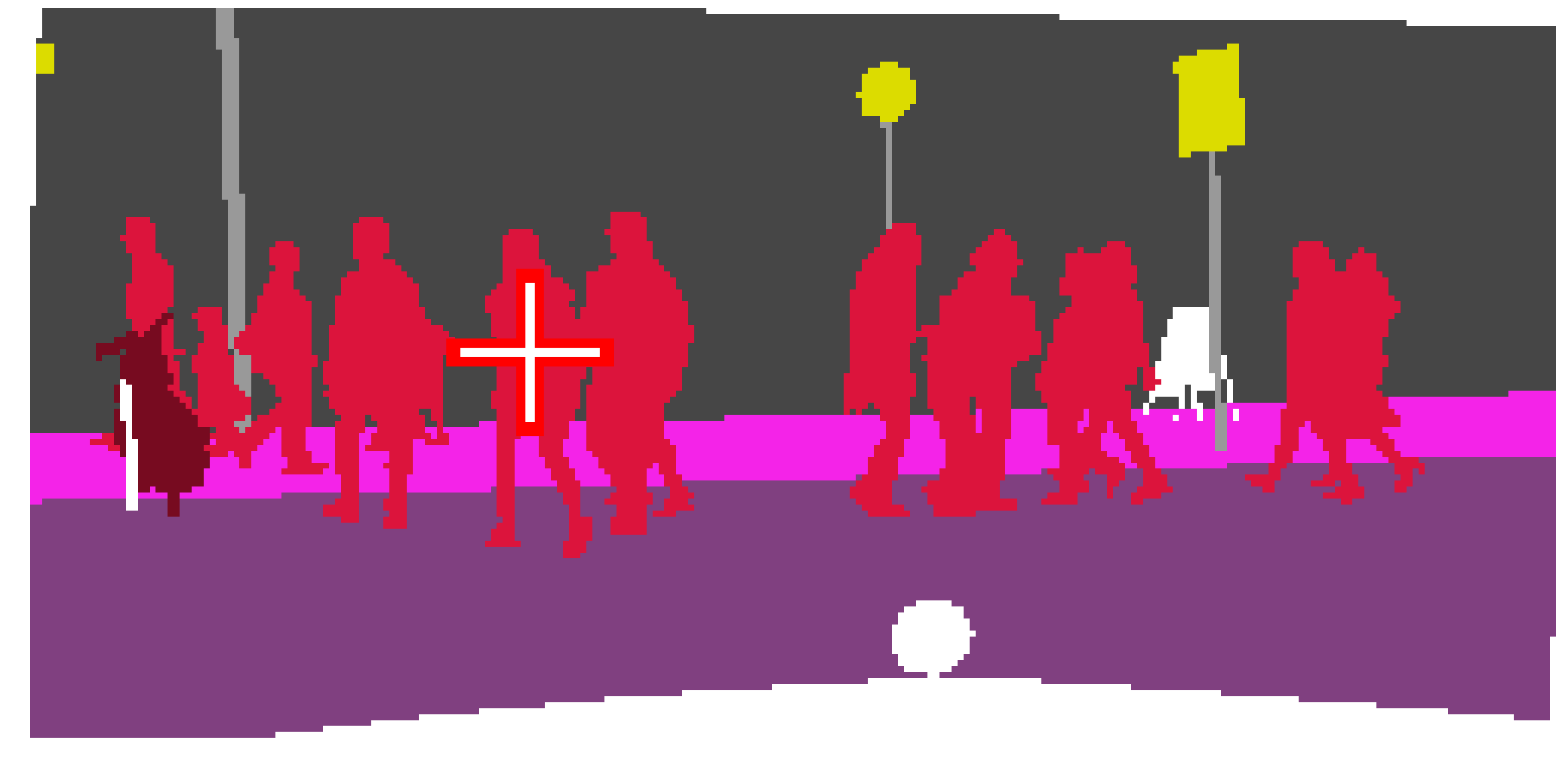}
		\includegraphics[width=.15\textwidth]{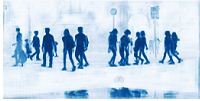}\\
		\vspace{.05cm}
		\includegraphics[width=.15\textwidth]{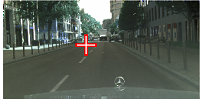}
		\includegraphics[width=.15\textwidth]{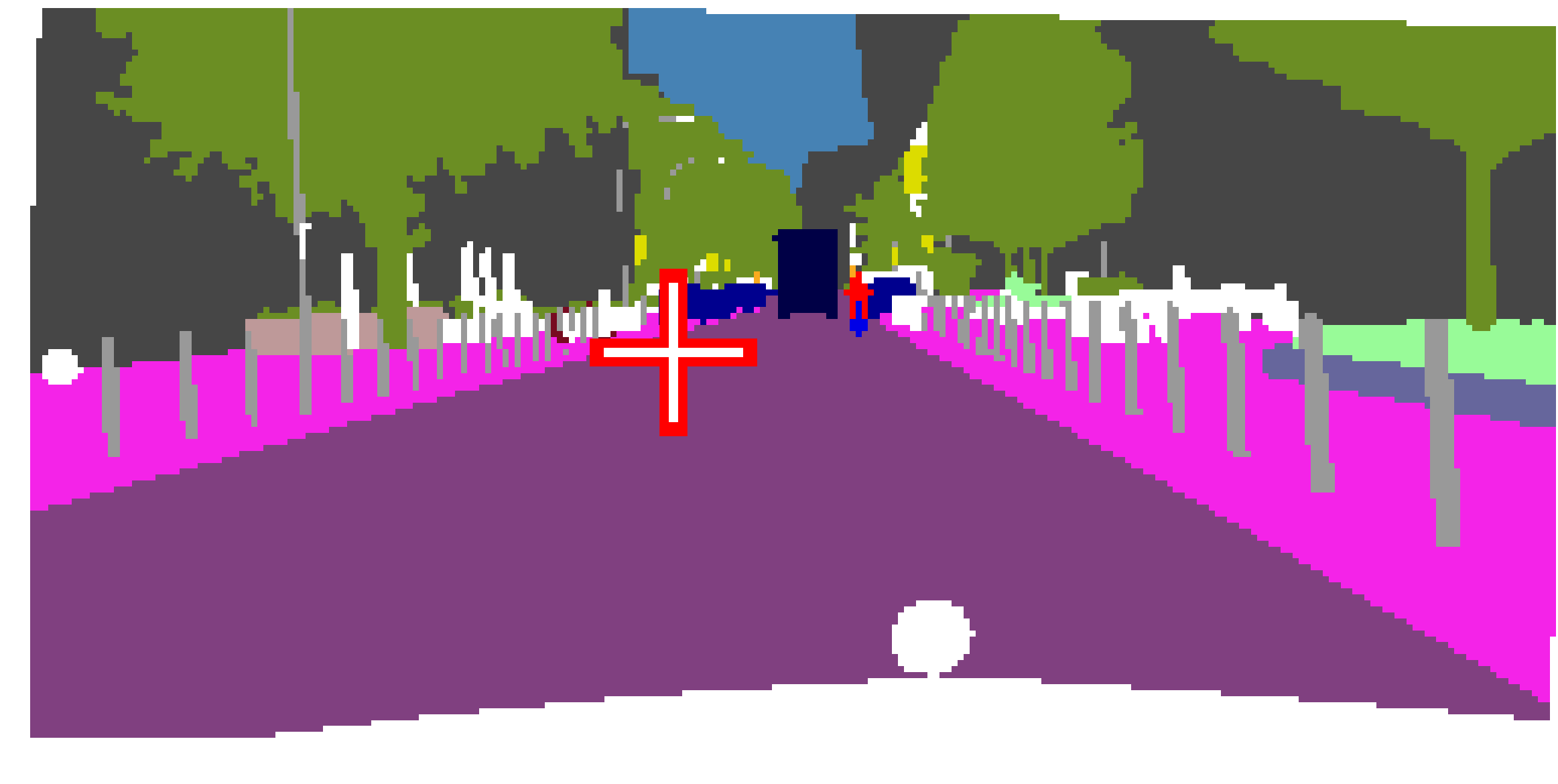}
		\includegraphics[width=.15\textwidth]{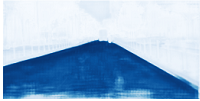}
		\caption{\small
			\textbf{Illustrating the predicted dense relation matrices.}
			The first column illustrates example images sampled from the Cityscapes \texttt{val},
			and we mark three pixels on object car, person and road with {\color{black}{\ding{57}}} respectively.
			The second column illustrates ground truth segmentation maps.
			The third column illustrates the dense relation matrices
			(or approximated object context maps) 
			of the three pixels.
			We can see that 
			the relation values
			corresponding to the pixels 
			belonging to the same category 
			as the selected pixel tend
			to be larger.
		}
		\label{fig:example_dense_relation}
	\end{figure}

    To explicitly emphasize the contribution of the object pixels,
	we present an object context that aims at only gathering the pixels that 
	belong to the same category as a given pixel as its context.
	Compared to the conventional multi-scale context schemes,
	our object context pays more attention 
	to the necessary object information.
	Although estimating the accurate object context is not an easy task,
	we empirically find that a coarse estimation of the object context
	already outperforms both PPM and ASPP schemes on various benchmarks.

	For a given pixel,
	we can use a binary vector to record pixels that
	belong to the same category as it with $1$ and $0$ otherwise.
	Thus, a binary relation matrix of $N\times N$ can be used to record 
	the pair-wise relations between any two of $N$ pixels.
	Since computing the binary relation matrix is intractable,
	we use a dense relation matrix to serve as a surrogate of it,
	in which each relation value is
	computed based on the high-level features' inner-product similarities.
	Therefore, the relation value of the semantically similar pixels 
	tend to be larger. 
	In our implementation, we use the conventional
	self-attention scheme~\citep{vaswani2017attention}
    to predict the dense relation matrix, 
    which requires $\mathcal{O}(N^2)$ computation complexity.
    To address the efficiency problem,
	we propose a new interlaced sparse self-attention scheme that
	significantly improves the efficiency while maintaining the performance 
	via two sparse relation matrices to approximate the dense relation matrix.
	To illustrate that our approach is capable of enhancing
	the object pixels, we show some examples of the 
	predicted dense relation matrices in Fig.~\ref{fig:example_dense_relation},
	where the relation values on the object pixels
	are larger than the relation values on the background pixels.
	
	We further illustrate two extensions that capture richer context information:
	(i) pyramid object context,
	which estimates the object context
	within each sub-region generated by the spatial pyramid partitions 
	following the PPM~\citep{zhao2017pyramid}. 
	(ii) atrous spatial pyramid object context,
	which combines ASPP~\citep{chen2017rethinking} with the object context.
	\textcolor{black}
	{
	We summarize our main contributions as following:
	\begin{itemize}
	\item We present a new object context scheme that
	explicitly enhances the object information.
	\item We propose to instantiate the object context scheme
	with an efficient interlaced sparse self-attention that
	significantly decreases the complexity compared to the conventional self-attention scheme.
	\item We construct the OCNet based on three kinds of object context modules
	and achieve competitive performance on five challenging semantic segmentation benchmarks including Cityscapes, ADE20K, LIP, PASCAL-Context and COCO-Stuff.
	\end{itemize}
	}
	
	\section{Related Work}

	\paragraph{Resolution.}
	Earlier studies based on conventional FCN~\citep{long2015fully} apply 
	the consecutive convolution striding and pooling operations
	to extract low-resolution feature map with high-level semantic information.
	For example, 
	the output feature map size of ResNet-$101$ is 
	$\frac{1}{32}$ of the input image,
	and such significant loss of the spatial information is one of the main challenges
	towards accurate semantic segmentation.
	To generate high-resolution feature map without much loss
	of the semantic information,
	many efforts~\citep{ronneberger2015u,badrinarayanan2017segnet,yu2015multi,chen2017rethinking,SunXLW19} have proposed various efficient mechanisms.
	In this paper,
	we adopt the dilated convolution~\citep{yu2015multi,chen2017rethinking} on ResNet-$101$
	to increase the output stride from $32$ to $8$ by 
	following the same settings of PSPNet~\citep{zhao2017pyramid}.
	Besides, we also conduct experiments based on 
	the recent HRNet~\citep{SunXLW19} with output stride $4$.
	We empirically verify that our approach
	is more efficient than the conventional multi-scale context mechanism, PPM and ASPP,
	with high resolution output feature map.
	More detailed comparisons are summarized in Table~\ref{table:efficiency_compare}.

	\vspace{.1cm}
	\paragraph{Context.}
	Context plays an important role in various computer vision tasks and it is of various forms such as global scene context, geometric context, relative location, 3D layout and so on. 
	Context has been investigated for both object detection~\citep{divvala2009empirical} and part detection~\citep{Gonzalez-Garcia_2018_CVPR}. 
	
	The importance of context for semantic segmentation is also verified in the recent works~\citep{liu2015parsenet,zhao2017pyramid,chen2017rethinking,shetty2019not}.
	It is common to define the context as a set of pixels in the literature of semantic segmentation.
	Especially, we can divide most of the existing context mechanisms into two kinds:
	(i) nearby spatial context: ParseNet~\citep{liu2015parsenet} treats all pixels over the whole image as the context, and PSPNet~\citep{zhao2017pyramid} performs pyramid pooling over sub-regions of four pyramid scales and all pixels within the same sub-region are treated as
	the context for the pixels belonging to the sub-region.
	(ii) sampled spatial context: DeepLabv3~\citep{chen2017rethinking} applies multiple atrous convolutions with different atrous rates to capture spatial pyramid context information and regards these spatially regularly sampled pixels as the context.
	
	These two kinds of context are defined over regular rectangle regions 
	and might carry pixels belonging to the background categories.
	Different from them, our object context is defined as the 
	set of pixels belonging to the same object category, 
	emphasizing the object pixels that are essential for labeling the pixel.
	There also exist some con-current efforts~\citep{fu2018dual,zhang2019co,huang2018ccnet,li2019ema,psanet} that exploit
	the semantic relations between pixels to construct the context,
	and our approach is different from most of them as we propose a simple yet effective interlaced sparse mechanism to
	model the relational context with smaller computation cost.
	
	\vspace{.1cm}
	\paragraph{Attention.}
	Self-attention~\citep{vaswani2017attention} and non-local neural network~\citep{wang2018non}
	have achieved great success on various tasks with its efficiency on modeling long-range
	contextual information.
	The self-attention scheme~\citep{vaswani2017attention} calculates the context at one position as a aggregation of all positions in a sentence {\color{black}(at the encoder stage)}.
	Wang \etal~ further proposed the non-local neural network~\citep{wang2018non} for vision tasks such as video classification, object detection and instance segmentation based on self-attention scheme.
	
	Our implementation is inspired by the self-attention scheme.
	We first apply the self-attention scheme 
	to predict the dense relation matrix
	and verify its capability to approximate the object context,
	and there also exist some concurrent studies~\citep{fu2018dual,zhang2019co}
	{\color{black}that apply the self-attention scheme for semantic segmentation.}
	Some recent efforts~\citep{huang2018ccnet,yue2018cgnl,zhu2019asymmetric}
	propose different mechanisms to decrease the computation complexity 
	and memory consumption of self-attention scheme.
	For example, CGNL~\citep{yue2018cgnl} (Compact Generalized Non-local) applies the Taylor series of the RBF kernel function to approximate the pair-wise similarities,
	RCCA~\citep{huang2018ccnet} (Recurrent Criss-Cross Attention) applies two consecutive criss-cross attention to approximate
	the original self-attention scheme.
	
	Our interlaced sparse self-attention scheme is different from both
	CGNL and RCCA through factorizing the dense relation matrix
	to two sparse relation matrices,
	and we find that similar mechanisms have been applied in the 
	previous studies on network architecture design including
	ShuffleNet~\citep{ma2018shufflenet} and
	Interleaved Group Convolution~\citep{igcv2,igcv1}.
	The concurrent sparse transformer~\citep{child2019generating}
	also apply the similar mechanism on one dimensional 
	text/audio related tasks that require sequential masked inputs.

	\section{Approach}
	\label{methods}
	We introduce our approach with four subsections.
	First, we introduce the general mathematical
	formulation of the context representation
	and the definition of object context (Sec.~\ref{formulation}).
	Second, we instantiate the object context with
	the conventional self-attention (SA)~\citep{vaswani2017attention} and
	our interlaced sparse self-attention (ISA) (Sec.~\ref{instantiations}).
	Third, we present the pyramid extensions of object context (Sec.~\ref{pyramid_object_context}).
	Last, we illustrate the overall pipeline and the implementation details of OCNet (Sec.~\ref{architecture}).

	\textbf{
		\subsection{Formulation}
		\label{formulation}
	}
	
	\vspace{.1cm}
	\paragraph{Preliminary.}
	We define the general mathematical formulation 
	of the context representation as:
	\begin{ceqn}
    \begin{align}
    \mathbf{z}_i = \rho(\frac{1}{|\mathcal{I}_i|}\sum_{j \in \mathcal{I}_i} \delta(\mathbf{x}_j)).
    \label{eq:general_context}
    \end{align}
	\end{ceqn}
	
	We use $\mathbf{X}$ and $\mathbf{Z}$ to represent
	the input representation and the context representation respectively.
	$\mathbf{x}_j$ is the $j$-th element of $\mathbf{X}$ 
	and $\mathbf{z}_i$ is the $i$-th element of $\mathbf{Z}$.
	$\delta(\cdot)$ and $\rho(\cdot)$ are two
	different transform functions.
	$\mathcal{I}=\{1,\cdots,N\}$ represents a set of $N$ pixels.
	We use $\mathcal{I}_i$ to represent a subset of $\mathcal{I}$,
	in other words, $\mathcal{I}_i$ is the set of context pixels for pixel $i$.
	We show how $\mathcal{I}_i$ selects pixels in the following discussions.
	Intuitively, the above formula of context representation is to describe a pixel with
	the weighted average representations of a set of relevant pixels.
	
	The above mathematical formulations are based on the one-dimensional case
	for convenience and can be generalized to the higher dimensional cases easily.
	One of the main differences between the existing representative context methods
    is the formulation of the context $\mathcal{I}_i$.

    \vspace{.1cm}
    \paragraph{Multi-scale context.}
    \textcolor{black}{
	\citet{zhao2017pyramid} proposes the pyramid pooling (PPM)
	context that constructs the $\mathcal{I}_i$ as a set of
	spatially-close pixels around pixel $i$ within
	the regular regions of different scales.
	\citet{chen2017rethinking} introduces the atrous  spatial  pyramid  pooling (ASPP)
	context that estimates $\mathcal{I}_i$ as a set of sparsely sampled pixels with 
	different dilation rates around pixel $i$.}
	
	\textcolor{black}{
	We take the representative multi-scale context scheme ASPP as an example
	to illustrate the formulation of $\mathcal{I}_i$:
    \begin{ceqn}
    \begin{align}
    \mathcal{I}_i = \bigcup_{r\in \{12, 24, 36\}} \{j\in\mathcal{I}~|~|i - j| = r\},
    \label{eq:aspp_context}
    \end{align}
    \end{ceqn}
    where $r\in \{12, 24, 36\}$ is the dilation rate,
    $|i - j|$ represents the spatial distances between pixel $i$ and $j$,
    and $\mathcal{I}_i$ is defined over one-dimensional input.
    Therefore, the ASPP context of pixel $i$ is a set of sampled pixels that have
    the predefined spatial distances with $i$.
	Besides, we illustrate the formulation of $\mathcal{I}_i$ 
	based on PPM scheme in Appendix A.}
	
	Both kinds of context tend to be a mixture of object pixels and 
	background pixels. Therefore, they have not explicitly 
	enhanced the contribution from the object pixels.
	Motivated by the fact that the category of each pixel
	is essentially inherited from the category of 
	the object that it lies in, we propose a new scheme
	named object context to
	explicitly enhance the information of object pixels.
    
	\vspace{.1cm}
	\paragraph{Object context.}

	We define the \emph{object context} for pixel $i$ as:
    \begin{ceqn}
    \begin{align}
    \mathcal{I}_i = \{j\in\mathcal{I} | l_j = l_i\},
    \label{eq:object_context}
    \end{align}
    \end{ceqn}
	where $l_i$ and $l_j$ are the label of pixel $i$ and $j$ respectively.
	We can see the object context for pixel $i$
	is essentially a set of pixels 
	that belong to the same object category as $i$.

	\textcolor{black}{
	We can represent the pairwise relations between any two of $N$ pixels (encoded in the ground-truth object context) with a 
	binary relation matrix of $N\times N$,
	where the $i$-th row records all pixels belonging to the same 
	category with pixel $i$ with $1$ and $0$ otherwise.
	Especially, the binary relation matrix only encodes partial 
	information of the ground-truth object context, i.e.,
	the (same category) label co-occurring relations. In other words, all classes could be permuted and the binary relation matrix would be unchanged.}
	
	Considering it is intractable to estimate the binary relation matrix,
	we propose to use a dense relation matrix to serve as a surrogate of the binary relation matrix. We expect that the relation values between the pixels belonging to the same object category are larger than the ones belonging to different categories,
	thus, the contributions of the object pixels
	are enhanced.
	
	In the following discussions,
	we first illustrate the formulation of the dense relation scheme that directly estimates the dense relation matrix $\mathbf{W}$ of size $N\times N$.
	Second, to improve efficiency,
	we propose a sparse relation scheme
	that factorizes the dense relation matrix as the combination of two sparse relation matrices including $\mathbf{W}^l$ and $\mathbf{W}^g$, where both sparse relation matrices are of size $N\times N$.
	More details are illustrated as follows.

    \vspace{.1cm}
	\paragraph{Dense relation.}
	The dense relation scheme estimates the relations 
	between each pixel $i$ and all pixels in $\mathcal{I}$.
	We illustrate the context representation based on dense relation:
	\begin{ceqn}
		\begin{align}
		\mathbf{z}_i = \rho(\sum_{j \in \mathcal{I}}
		w_{ij} \delta(\mathbf{x}_j)),
		\label{eq:dense_relation}
		\end{align}
	\end{ceqn}
	where $w_{ij}$ is the relation value between
	pixel $i$ and $j$, i.e., the element of $\mathbf{W}$ at coordinates ($i$, $j$).
	As we need to estimate the relations between $i$ and 
	all pixels in $\mathcal{I}$ directly,
	the computational complexity of estimating $\mathbf{W}$
	is quadratic to the input size: $\mathcal{O}(N^2)$.
	
	\paragraph{Sparse relation.}
	The sparse relation scheme only estimates the relations
	between the pixel $i$ and two subsets of selected pixels
	following the ``interlacing (a.k.a. interleaving) method''~\citep{greenspun1999philip,roelofs1999png}.
	We illustrate the context representation based on sparse relation:
	\begin{ceqn}
		\begin{align}
		\mathbf{z}^g_i = \rho(\sum_{j \in \mathcal{I}_{i}^{g}} 
		w^{g}_{ij} \delta(\mathbf{x}_j)),\\
		\mathbf{z}^l_i = \rho(\sum_{j \in \mathcal{I}_{i}^{l}}
		w^{l}_{ij} \delta(\mathbf{z}^g_j)).
		\label{eq:sparse_relation}
		\end{align}
	\end{ceqn}

	We use the superscript $g$ / $l$ to mark the operators and operations
	associated with the global / local relation stage respectively.
	For example, $\mathbf{z}^g_i$ / $\mathbf{z}^l_i$ represents the
	context representation after the global / local relation stage of the $i$-th pixel.
	Refer to Sec.~\ref{instantiations} for more details.
	$w^{g}_{ij}$ / $w^{l}_{ij}$ is the relation
	between pixel $i$ and pixel $j$ that belongs to
	$\mathcal{I}_{i}^{g}$ / $\mathcal{I}_{i}^{l}$ respectively.
	$\mathcal{I}_{i}^{g}$ and $\mathcal{I}_{i}^{l}$ are the selected context pixels:
	\begin{ceqn}
	\begin{align}
	\mathcal{I}_{i}^{g}&=\{j\in\mathcal{I} : j \equiv i\pmod{P}\}
	\label{eq:global_index}, \\
    \mathcal{I}_{i}^{l}&=\{j\in\mathcal{I} : \lfloor\frac{j-1}{P}\rfloor = \lfloor\frac{i-1}{P}\rfloor\},
	\label{eq:local_index}
	\end{align}
	\end{ceqn}
	where $\mathcal{I}_{i}^{g}$ / $\mathcal{I}_{i}^{l}$ is a subset of pixels with the same remainder / quotient as the pixel $i$ when divided by $P$ respectively.
	Both $i$ and $j$ in the above illustrations represent
	the spatial positions of pixel $i$ and $j$ in the one-dimensional case.
	$P$ represents the group number in the global relation stage (Sec.~\ref{instantiations}) and it determines the selection of context pixels.
	The main advantage of the sparse relation scheme lies at
	we only need to estimate the relation values
	between pixel $i$ and $\mathcal{I}_{i}^{g} \cup \mathcal{I}_{i}^{l}$ instead of $\mathcal{I}$, thus, saves a lot of computation cost.

    \begin{figure*}[t]
		\includegraphics[width=\textwidth]{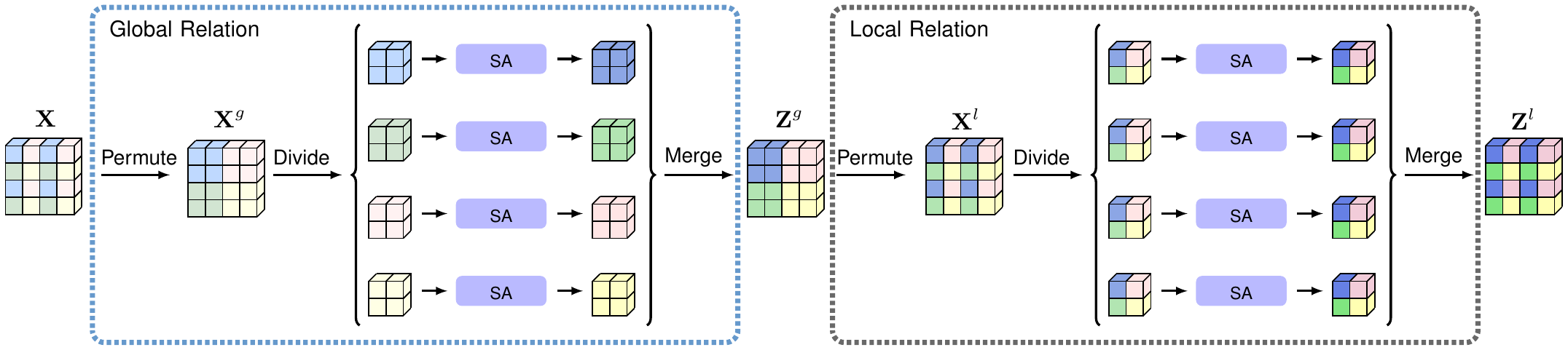}
		\vspace{-5mm}
		\caption{\small{
				\textbf{Illustrating the Interlaced Sparse Self-Attention}.
				Our approach is consisted of a 
				global relation module and a local relation module.
				The feature map in the left-most/right-most is the input/output.
				First, we color the input feature map $\mathbf{X}$ with four different colors.
				We can see that there are $4$ local groups and each group is
				consisted of four different colors.
				For the global relation module, we permute and group (divide) all positions 
				with the same color having long spatial interval distances together in $\mathbf{X}$,
				which outputs $\mathbf{X}^{g}$.
				Then, we divide the $\mathbf{X}^{g}$ into $4$ groups and apply the self-attention 
				on each group independently.
				We merge all the updated feature map of each group together 
				as the output $\mathbf{Z}^{g}$.
				For the local relation module, we permute the $\mathbf{Z}^{g}$
				to group the originally nearby 
				positions together and get $\mathbf{X}^{l}$.
				Then we divide and apply self-attention following the same manner as global relation,
				obtain the final feature map $\mathbf{Z}^{l}$.
				We can propagate the information from all input positions to
				each output position with the combination of the global relation module and the local relation module.
				{\color{black}{
				The positions with the same saturation of color mark the value of the feature maps are kept unchanged. We only increase the saturation the color when we update the feature maps with self-attention operation.}}
		}}
		\label{fig:interlaced_attn}
	\end{figure*}

    {\textcolor{black}{
	Considering the pixels with equal remainder / quotient 
	share the same $\mathcal{I}_{*}^{g}$ / $\mathcal{I}_{*}^{l}$,
	thus, we ignore the subscript and use $\mathcal{I}^{g}$ / $\mathcal{I}^{l}$ to
	represent all groups of pixels that share equal remainder / quotient respectively.
	We compute (part of) the sparse relation matrices $\mathbf{W}^g$ / $\mathbf{W}^l$
	within each group of pixels in $\mathcal{I}^{g}$ / $\mathcal{I}^{l}$ respectively.
	Especially, both $\mathbf{W}^g$ and $\mathbf{W}^l$
	are sparse block matrices
	and we can approximate the original dense relation
	matrix as the product of these two sparse
	relation matrices:
	\begin{ceqn}
	\begin{align}
	\mathbf{W} = \mathbf{W}^l \mathbf{P}^{\top} \mathbf{W}^g \mathbf{P},
	\label{eq:product_of_sparse_matrix}
	\end{align}
	\end{ceqn}
    where we use $\mathbf{P}$ to represent a 
	permutation matrix of size $N \times N$
	that ensures the pixel orderings of the two sparse relation matrices are matched
	and $\mathbf{P}^{\top}$ is the transpose of $\mathbf{P}$.
	We illustrate the definition of each value $p_{i,j}$ in $\mathbf{P}$ in Appendix B
	and why the sparse relation scheme is more efficient than the dense relation scheme in Appendix C.
    }}

    \textbf{
		\subsection{Instantiations}
		\label{instantiations}
	}
	We explain the specific instantiations of
	dense relation and sparse relation 
	based on the self-attention
	and the interlaced sparse self-attention respectively.
	
	\vspace{0.5mm}
	\paragraph{Self-attention.}
	The implementation of dense relation scheme based on self-attention is 
	illustrated as following,
	\begin{ceqn}
    \begin{align}
    \mathbf{W} &= \mathrm{Softmax} (\frac{\theta(\mathbf{X})\phi(\mathbf{X})^{\top}}{\sqrt{d}}), \label{eq:sa1}\\
    \mathbf{Z} &= \rho(\mathbf{W} \delta(\mathbf{X})), \label{eq:sa2}
    \end{align}
	\end{ceqn}
	
	$\mathbf{X}\in \mathbb{R}^{ N\times C_{\rm{in}}}$ is the input representation,
	$\mathbf{W}\in \mathbb{R}^{N\times N}$ is the dense relation matrix,
	and $\mathbf{Z}\in \mathbb{R}^{N\times C_{\rm{out}}}$ is the output representation.
	We assume ${C_{\rm{in}}} = {C_{\rm{out}}} = C$ in the following discussion for convenience.
	$\theta$ and $\phi$ are two different functions that
	transform the input to lower dimensional space and 
	$\theta(\mathbf{X}), \phi(\mathbf{X}) \in \mathbb{R}^{N\times \frac{C}{2}}$.
	The inner product in the lower dimensional space is used to compute
	the dense relation matrix $\mathbf{W}$.
	The scaling factor $d$ is used to
	to solve the small gradient problem of softmax function 
	according to~\citet{vaswani2017attention} and we set $d=\frac{C}{2}$.
	Self-attention uses the function $\rho$ and $\delta$ to learn a better embedding 
	and we have $\rho(\cdot) \in \mathbb{R}^{N\times C}$ and $\delta(\cdot) \in \mathbb{R}^{N\times \frac{C}{2}}$.
	{\color{black}{
	According to the original description of self-attention in ~\citet{vaswani2017attention}, 
	we can also call $\theta$, $\phi$ and $\delta$ as query-, key-, and value-transform function respectively.
	}}
	
	We implement both 
	$\theta(\cdot)$ and $~\phi(\cdot)$
	with {\color{black}{two consecutive groups of}} $1\times 1~\operatorname{conv} \rightarrow \operatorname{BN} \rightarrow \operatorname{ReLU}$.
	\textcolor{black}{BN is the abbreviation for batch normalization~\citep{ioffe2015batch}
	that synchronize the statistics.}
	We implement 
	$\delta(\cdot)$ and $~\rho(\cdot)$
	with $1\times 1~\operatorname{conv}$.
	Specifically, $\theta(\cdot)$, $~\phi(\cdot)$ and $\delta(\cdot)$
	halve the input channels while $~\rho(\cdot)$
    doubles the input channels.
    
	\vspace{.2cm}
	\paragraph{Interlaced sparse self-attention.}
	The implementation of the sparse relation scheme, 
	i.e., the interlaced sparse self-attention, 
	first divides all pixels into multiple subgroups
	and then applies the self-attention
	on each subgroup to compute the sparse relation matrices,
	i.e., $\mathbf{W}^{g}$ and $\mathbf{W}^{l}$, and the
	context representations.
	
	We illustrate the overall pipeline of 
	the interlaced sparse self-attention scheme
	with a two dimensional example in Fig.~\ref{fig:interlaced_attn},
	where we estimate $\mathbf{W}^g$ 
	with the global relation module
	and $\mathbf{W}^l$ with the local relation module.
	With the combination of these two sparse relation matrices,
	we can approximate the 
	dense relations between any two of all pixels,
	which is explained with an example in Appendix D.

    \vspace{.2cm}
	\noindent\emph{Global relation}.
	{\color{black}{
	We divide all positions into multiple groups
	with each group consists of a subset of sampled positions
	according to the definition of $\mathcal{I}^{g}$.
	Considering that the pixels within each group are sampled
	based on the remainder divided by 
	the number of groups $P$ and they are distributed across
	the global image range, 
	thus, we call it global relation.
	
	We first permute the input feature map $\mathbf{X}$:
	\begin{ceqn}
    \begin{align}
    \mathbf{X}^{g} = \operatorname{Permute}(\mathbf{X}) = \mathbf{P} \mathbf{X}, \label{eq:permute_global}
    \end{align}
	\end{ceqn}	
	Second, we divide $\mathbf{X}^{g}$ into ${P}$ groups
	with each group containing ${Q}$ neighboring positions ($N={P} \times {Q}$):
	\begin{ceqn}
    \begin{align}
    \mathbf{X}^{g} \xrightarrow{\rm{Divide}} \{\mathbf{X}^{g}_1, \mathbf{X}^{g}_2, \cdots, \mathbf{X}^{g}_{P}\},  \label{eq:divide_global}
    \end{align}
	\end{ceqn}
	where each $\mathbf{X}^{g}_p \in \mathbb{R}^{ Q\times C}$ is 
	a subset of $\mathbf{X}^{g}$ and we have $\mathbf{X}^{g} = [{\mathbf{X}^{g}_1}^{\top}, {\mathbf{X}^{g}_2}^{\top}, \cdots, {\mathbf{X}^{g}_{P}}^{\top}]^{\top}$.
	}}
	Third, we apply the self-attention on each $\mathbf{X}^{g}_p$ independently:
	\begin{ceqn}
		\begin{align}
		\mathbf{W}_p^{g} &= \mathrm{Softmax} (\frac{\theta(\mathbf{X}^{g}_p)\phi(\mathbf{X}^{g}_p)^{\top}}{\sqrt{d}}), \label{eq:isa1}\\
		\mathbf{Z}^{g}_p &= \rho(\mathbf{W}_p^{g} \delta(\mathbf{X}^{g}_p)), \label{eq:isa2}
		\end{align}
	\end{ceqn}
	where $\mathbf{W}_p^{g} \in \mathbb{R}^{{Q}\times {Q}}$
	is a small dense relation matrix based on all positions
	from $\mathbf{X}_p^{g}$, 
	$\mathbf{Z}^{g}_p \in \mathbb{R}^{{Q}\times C}$ is the updated representation
	based on $\mathbf{X}^{g}_p$, and $d$ takes the same value as in the previous Equation~\ref{eq:sa1}.
	We apply the same implementation for
	all transform functions including
	$\theta(\cdot)$, $~\phi(\cdot)$, 
	$\delta(\cdot)$ and $~\rho(\cdot)$
	following the implementation of self-attention.
	We illustrate the overall sparse relation matrix $\mathbf{W}^{g}$ in the global relation stage:
	\begin{ceqn}
		\begin{align}
		\mathbf{W}^{g} & = 
		\begin{bmatrix}
		\mathbf{W}^{g}_{1} & 0 & \cdots & 0 \\
		0 & \mathbf{W}^{g}_{2} & \cdots & 0 \\
		\vdots     &\vdots      & \ddots & \vdots \\
        0 & 0 & \cdots & \mathbf{W}^{g}_{P}
        \label{eq:isa3}
        \end{bmatrix}, 
		\end{align}
	\end{ceqn}
	where only the relation values in the diagonal blocks are non-zero.
	Therefore, we only need to estimate the relation values between
	pixel pairs belonging to the same group and
	ignore the relations between pixel pairs from different groups.
	
	{\color{black}{
	We {\color{black}concatenate} all $\mathbf{Z}^{g}_p$ from different groups
	and get the output representation $\mathbf{Z}^{g} = [{\mathbf{Z}^{g}_1}^{\top}, {\mathbf{Z}^{g}_2}^{\top}, \cdots, {\mathbf{Z}^{g}_{P}}^{\top}]^{\top}$ after the global relation stage:
	\begin{ceqn}
    \begin{align}
    \{\mathbf{Z}^{g}_1, \mathbf{Z}^{g}_2, \cdots, \mathbf{Z}^{g}_{P}\} \xrightarrow{\rm{Merge}} \mathbf{Z}^{g},  \label{eq:concate_global}
    \end{align}
	\end{ceqn}
	}}
	
	\vspace{.2cm}
	\noindent\emph{Local relation}.
	{\color{black}{
	In the local relation stage,
	we divide the positions into multiple groups
	according to the definition of $\mathcal{I}^{l}$,
	where each group of pixels are sampled
	based on the quotient and they are distributed within the 
	local neighboring range,
	thus, we call it local relation.
	
	We apply another permutation on the output feature map from
	the global relation module following:
	\begin{ceqn}
    \begin{align}
    \mathbf{X}^{l} = \operatorname{Permute}(\mathbf{Z}^{g}) = \mathbf{P}^{\top}\mathbf{Z}^{g}, \label{eq:permute_local}
    \end{align}
	\end{ceqn}
	Then, we divide $\mathbf{X}^{l}$ into ${Q}$ groups
	with each group containing ${P}$ neighboring positions:
	\begin{ceqn}
    \begin{align}
    \mathbf{X}^{l} \xrightarrow{\rm{Divide}} [{\mathbf{X}^{l}_1}, {\mathbf{X}^{l}_2}, \cdots, {\mathbf{X}^{l}_{Q}}], \label{eq:divide_local}
    \end{align}
	\end{ceqn}
	where each $\mathbf{X}^{l}_q \in \mathbb{R}^{{P}\times C}$ and we have
	$\mathbf{X}^{l} = [{\mathbf{X}^{l}_1}^{\top}, {\mathbf{X}^{l}_2}^{\top}, \cdots, {\mathbf{X}^{l}_{Q}}^{\top}]^{\top}$.}}
	
	We apply the self-attention on each $\mathbf{X}^{l}_q$ independently,
	which is similar with the Equation~\ref{eq:isa1} and Equation~\ref{eq:isa2}
	in the global relation module.
	Accordingly, we can get $\mathbf{W}_q^{l}$ and $\mathbf{Z}_q^{l}$,
	where $\mathbf{W}_q^{l} \in \mathbb{R}^{{P}\times {P}}$
	is a small dense relation matrix based on $\mathbf{X}_q^{l}$,
	$\mathbf{Z}^{l}_q \in \mathbb{R}^{{P}\times C}$
	is the updated representation based on $\mathbf{X}_q^{l}$.
	We illustrate the sparse relation matrix computed based on
	the local relation:
	\begin{ceqn}
		\begin{align}
		\mathbf{W}^{l} & = 
		\begin{bmatrix}
		\mathbf{W}^{l}_{1} & 0 & \cdots & 0 \\
		0 & \mathbf{W}^{l}_{2} & \cdots & 0 \\
		\vdots     &\vdots      & \ddots & \vdots \\
		0 & 0 & \cdots & \mathbf{W}^{l}_{Q}
		\end{bmatrix},  
		\end{align}
	\end{ceqn}
	where the above relation matrix $\mathbf{W}^{l}$
	based on the local relation is also very sparse 
	and most of the relation values are zero.
	
	{\color{black}{
	We concatenate all $\mathbf{Z}^{l}_q$ from different groups
	and get output representation $\mathbf{Z}^{l} = [{\mathbf{Z}^{l}_1}^{\top}, {\mathbf{Z}^{l}_2}^{\top}, \cdots, {\mathbf{Z}^{l}_{Q}}^{\top}]^{\top}$ after the local relation stage:
	\begin{ceqn}
    \begin{align}
    \{\mathbf{Z}^{l}_1, \mathbf{Z}^{l}_2, \cdots, \mathbf{Z}^{l}_{Q}\} \xrightarrow{\rm{Merge}} \mathbf{Z}^{l},  \label{eq:concate_local}
    \end{align}
	\end{ceqn}
	where $\mathbf{Z}^{l}$ is also the final output representation of interlaced sparse self-attention scheme.
	}}
		
	\vspace{.2cm}
	\noindent\textbf{Complexity}.
	Given an input feature map of size $H\times W  \times C$,
	we analyze the computation/memory cost of both the self-attention mechanism
	and our interlaced sparse self-attention scheme
    as follows.

\begin{figure}[t]
		\includegraphics[width=.475\textwidth]{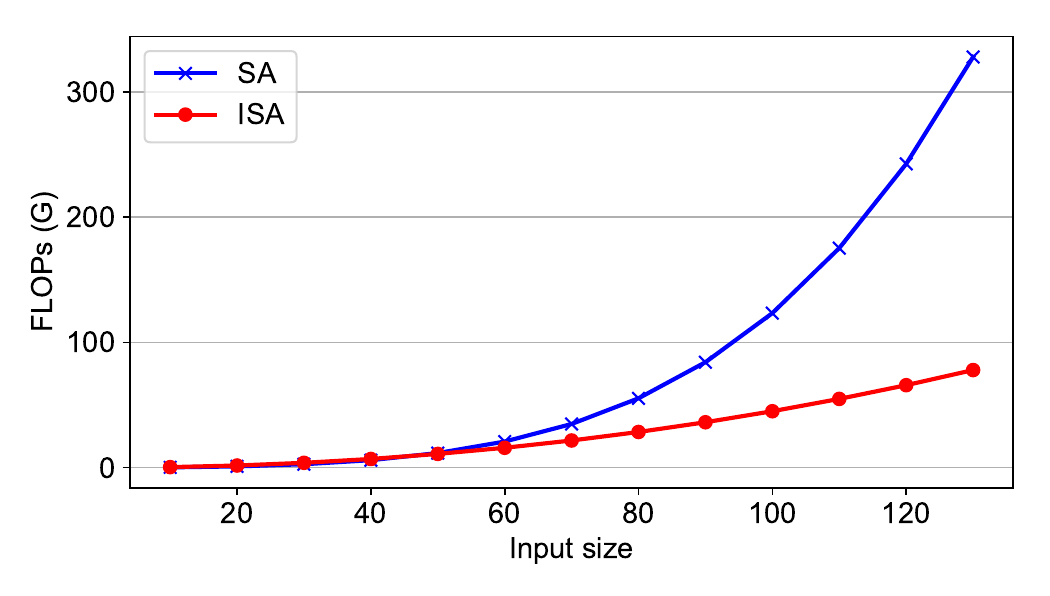}
		\caption{\small{\textbf{FLOPs \emph{vs.} input size}. 
				The $x$-axis represents the height or width of the input feature map (we assume
				the height is equal to the width for convenience)
				and the $y$-axis represents the computation cost measured with GFLOPs.
				We can see that the GFLOPs of self-attention (SA) mechanism
				increases much faster than our interlaced sparse self-attention (ISA) mechanism with inputs of higher resolution.
		}}
		\label{fig:complexity}
	\end{figure}

\begin{figure}[t]
		\includegraphics[width=.475\textwidth]{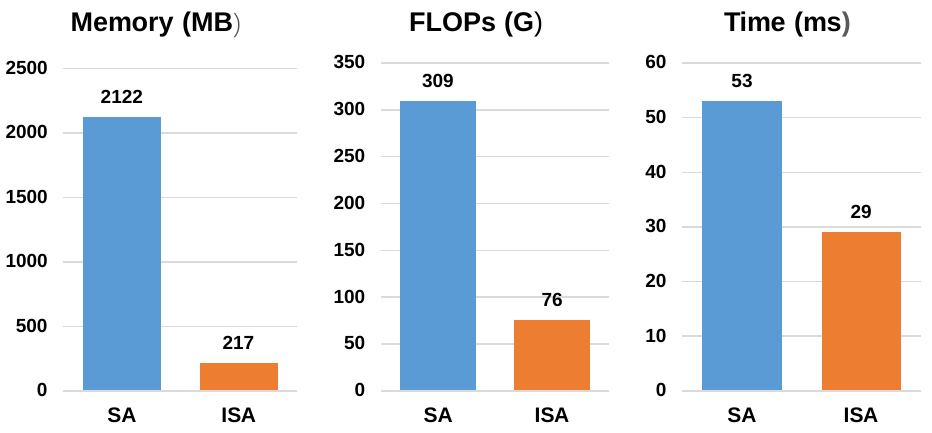}
		\caption{\small{\textbf{GPU memory/FLOPs/Running time comparison between SA and ISA}.
				All numbers are tested on a single Titan XP GPU with CUDA$8.0$ and
				an input feature map of $1\times 512\times 128\times 128$
				during inference stage.
				The lower, the better for all metrics.
				We can see that the proposed interlaced sparse self-attention (ISA) only uses $10.2\%$ GPU memory and $24.6\%$ FLOPs while being nearly
				$2\times$ faster when compared with the self-attention (SA).
		}}
		\label{fig:gpu_test_cost}
	\end{figure}

    {\color{black}{
	The computation complexity of self-attention mechanism is 
	$\mathcal{O}(HWC^2+(HW)^2{C})$,
	and the complexity of our interlaced sparse self-attention mechanism is
	$\mathcal{O}(HWC^2+(HW)^2{C}(\frac{1}{{P}_{h}{P}_{w}}+\frac{1}{{Q}_{h}{Q}_{w}}))$,
	where we divide the height dimension 
	into ${P}_{h}$ groups
	and the width dimension to ${P}_{w}$ groups 
	in the global relation stage
	and ${Q}_{h}$ and ${Q}_{w}$ groups
	during the local relation stage.
	We have $H = {P}_{h}{Q}_{h}$, $W = {P}_{w}{Q}_{w}$.
	The complexity of our approach can be {\color{black}{reduced to}} $\mathcal{O}(HWC^2+(HW)^{\frac{3}{2}}{C})$
	when ${P}_{h}{P}_{w} = \sqrt{HW}$.
	Detailed formulations and proof of the computation
	complexity are provided in Appendix E.
	}}
	We compare the theoretical GFLOPs of the interlaced sparse self-attention scheme 
	and the conventional self-attention scheme in Fig.~\ref{fig:complexity},
	where we can see that our interlaced sparse self-attention is much more 
	efficient than the conventional self-attention when processing inputs
	of higher resolution.
	We further report the actual 
	GPU memory cost (measured by MB),
	computation cost (measured by GFLOPs), 
	and inference time (measured by ms) 
	of both mechanisms 
	in Fig.~\ref{fig:gpu_test_cost} to illustrate the advantage of our method.	
	
	We present the PyTorch code of the
	proposed interlaced sparse 
	self-attention in Algorithm~\ref{fig:code}.
	{\color{black}{
	We explain the rough correspondence between 
	Fig.~\ref{fig:interlaced_attn} and  Algorithm~\ref{fig:code}.
	For example,
	in global relation stage,
	the combination of $\operatorname{reshape}$ in line-9 and 
	$\operatorname{permute}$ in line-10 of Algorithm~\ref{fig:code}
	corresponds to the $\operatorname{Permute}$ in Fig.~\ref{fig:interlaced_attn},
	and the $\operatorname{reshape}$ in line-11 of Algorithm~\ref{fig:code} corresponds to the 
	$\operatorname{Divide}$ in Fig.~\ref{fig:interlaced_attn}.
	Our implementation is optimized for the efficiency
	as we are applying the interlacing operations on the tensors of high dimension.
	Especially, the \texttt{permute} function (in both line-10 and line-16 of Algorithm~\ref{fig:code}) does not correspond to the \texttt{Permute} 
	(for one-dimensional situation)
	illustrated in Fig~\ref{fig:interlaced_attn}.
	}}

	{
		\lstset{
			backgroundcolor=\color{white},
			basicstyle=\fontsize{7.5pt}{8.5pt}\fontfamily{lmtt}\selectfont,
			columns=fullflexible,
			captionpos=b,
			commentstyle=\fontsize{8pt}{9pt}\color{codegray},
			keywordstyle=\fontsize{8pt}{9pt}\color{codegreen},
			stringstyle=\fontsize{8pt}{9pt}\color{codeblue},
			frame=none,
			otherkeywords = {self},
			autogobble=true,
			breaklines=true,
            numbers=left,
            stepnumber=1,    
            firstnumber=1,
            numberfirstline=true
		}
		\begin{algorithm}
			\tiny
			\begin{lstlisting}[language=python]
			def InterlacedSparseSelfAttention(x, P_h, P_w):
			# x: input features with shape [N,C,H,W]
			# P_h, P_w: Number of groups along H and W dimension
			
			N, C, H, W = x.size()
			Q_h, Q_w = H // P_h, W // P_w
			
			# global relation
			x = x.reshape(N, C, Q_h, P_h, Q_w, P_w)
			x = x.permute(0, 3, 5, 1, 2, 4)
			x = x.reshape(N * P_h * P_w, C, Q_h, Q_w)
			x = SelfAttention(x)
			
			# local relation
			x = x.reshape(N, P_h, P_w, C, Q_h, Q_w)
			x = x.permute(0, 4, 5, 3, 1, 2)
			x = x.reshape(N * Q_h * Q_w, C, P_h, P_w)
			x = SelfAttention(x)
			
			x = x.reshape(N, Q_h, Q_w, C, P_h, P_w)
			return x.permute(0, 3, 1, 4, 2, 5).reshape(N, C, H, W)
			\end{lstlisting}
			\vspace{-3mm}
			\caption{\small{Interlaced Sparse Self-Attention.}}
			\label{fig:code}
		\end{algorithm}
	}
	
\begin{figure*}[t]
		\centering
		\includegraphics[width=1\textwidth]{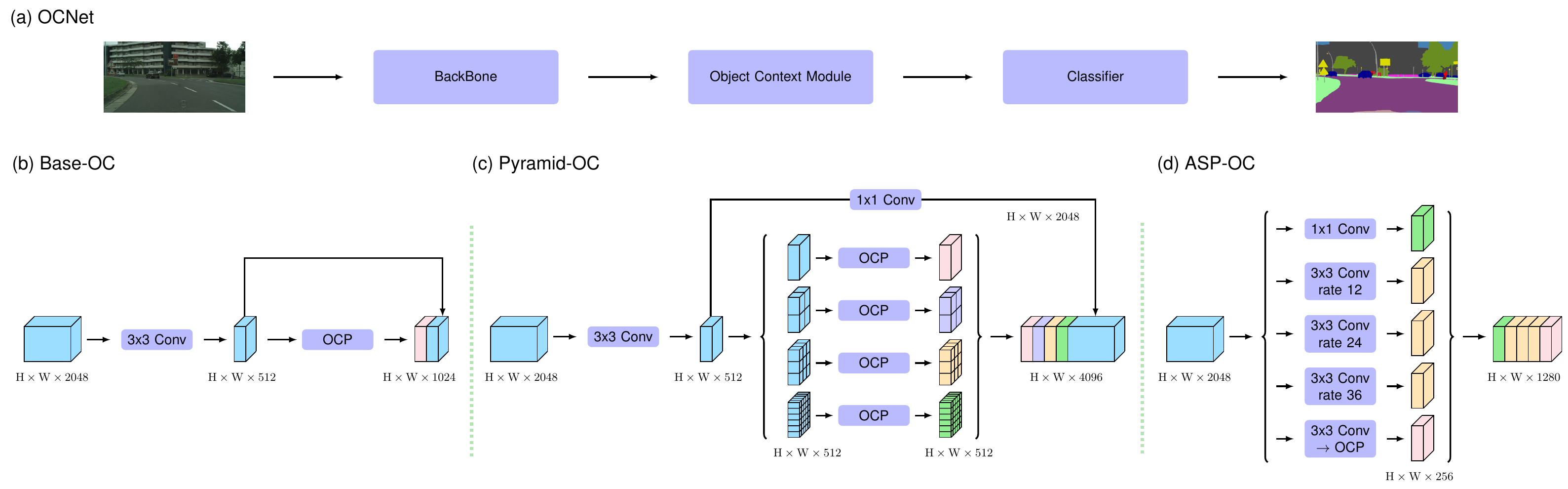}
		\vspace{-2mm}
		\caption{
			\small{
				\textbf{Illustrating the overall framework of OCNet}.
				(a) The overall network pipeline of OCNet: given an input image, we use a backbone to extract the
				feature map, then we apply an object context module on the feature map and output the contextual feature map. 
				Based on the contextual feature map, we apply a classifier to predict the final segmentation map.
				(b) Base-OC: we perform an object context pooling (OCP) on the input feature map, then we concatenate the
				output of OCP and the input feature map as the final output. 
				(c) Pyramid-OC: we apply four parallel OCPs independently. Each branch divides the input to different pyramid scales, 
				and the object context pooling is shared within each branch, then we concatenate the four output feature maps with a new feature map that is generated by {\color{black}{increasing the channels of the input feature map by $4\times$ ($512\to2048$)}}. 
				(d) ASP-OC: we apply an OCP and four dilated convolutions (these four branches
				are the same with the original ASPP and 
				{\color{black}{the rate represents the dilation rate}}), then we concatenate the five output feature maps as the output. {\color{black}{All the convolutions are followed by a group of $\operatorname{BN} \rightarrow \operatorname{ReLU}$ operation.}}
		}}
		\label{fig:ocnet_pipeline}
	\end{figure*}

	\vspace{.2cm}
	\noindent\textbf{Object context pooling}.
	We use self-attention or interlaced sparse self-attention
	to implement the object context pooling module (OCP).
	{
	\color{black}{
	The object context pooling estimates the context representation of each pixel $i$
	by aggregating the representations of the selected subset of pixels  based on the estimated dense relation matrix or the two sparse relation matrices.
	}
	}
	We first apply the object context pooling module based on either
	self-attention scheme or the proposed interlaced sparse self-attention scheme
	to compute the context representation,
	and then we \textcolor{black}{concatenate the input representation with 
	the context representation as the output representation},
	resulting in a baseline method named
	as \emph{Base-OC},
	and we illustrate the details in Fig.~\ref{fig:ocnet_pipeline}~(b).

	\textbf{
		\subsection{Pyramid Extensions}
		\label{pyramid_object_context}
	}
	To handle objects 
	of multiple scales\footnote{
    There exist two kinds of multiple scales problem: 
    (i) objects of different categories have multiple scales given their distances to the camera are the same, e.g., the ``car'' is larger than the ``person''. 
    (ii) the objects of the same category have multiple scales given their distances to the camera are different, e.g., the closer ``person'' is larger than the distance ``person''.
    },
	we further combine our approach
	with the conventional 
	multi-scale context schemes
	including PPM and ASPP.

    \vspace{2mm}
	\noindent\textbf{Combination with PPM}.
	Inspired by the previous pyramid pooling module~\citep{zhao2017pyramid},
	we divide the input image into regions of four pyramid scales:
	$1\times 1$ region, $2\times 2$ regions, $3\times 3$ regions and $6\times 6$ regions,
	and we update the feature maps for each
	scale by feeding the feature map of each region 
	into the object context pooling module respectively, 
	then we combine the four
	updated feature maps together. 
	Finally, we concatenate the multiple pyramid object context representations 
	with the input feature map. 
	We call the resulting
	method as \emph{Pyramid-OC}.
	More details are illustrated in Fig.~\ref{fig:ocnet_pipeline}~(c).
	
    \vspace{2mm}
	\noindent\textbf{Combination with ASPP}.
	The conventional atrous spatial pyramid pooling~\citep{chen2017rethinking}
	consists of $5$ branches including: an image-level pooling branch,
	a $1\times 1$ convolution branch and three $3\times 3$ dilated convolution branches with dilation rates being $12$, $24$ and $36$, respectively.
	We replace the image-level pooling branch with the object context pooling
	to exploit the relation-based object context information, resulting in a method
	which we name as \emph{ASP-OC}.
	More details are illustrated in Fig.~\ref{fig:ocnet_pipeline}~(d).

	\vspace{.1cm}
	\textbf{
		\subsection{Network Architecture}
		\label{architecture}
	}
	We illustrate the overall pipeline of 
	our OCNet in Fig.~\ref{fig:ocnet_pipeline}~(a).
	More details are illustrated as follows.

	\vspace{.1cm}
	\noindent\textbf{Backbone}.
	We use the ResNet-$101$~\citep{he2016deep} or HRNetV2-W$48$~\citep{SunZJCXLMWLW19}
	pretrained over the ImageNet dataset 
	as the backbone.
	For the ResNet-$101$,
	we make some modifications by following PSPNet~\citep{zhao2017pyramid}:
	replace the convolutions within the last two blocks 
	by dilated convolutions with dilation rates being $2$ and $4$, respectively,
	so that the output stride becomes $8$.
	For the HRNetV2-W$48$,
	we directly apply our approach on
	the final concatenated feature map with
	output stride $4$.
	
	\vspace{.1cm}
	\noindent\textbf{Base-OC}.
	Before feeding the feature map into the OCP,
	we apply a dimension reduction module (a $3 \times 3$ convolution)
	to reduce the channels of the feature maps output from the backbone 
	to $512$ for both ResNet-$101$ and HRNetV2-W$48$.
	Then we feed the updated feature map into the 
	OCP and concatenate the output feature map of the OCP with the 
	input feature map to the OCP.
	We further perform a $1 \times 1$ convolution 
	to decrease the channels of
	the concatenated feature map 
	from $1024$ to $512$, which is not included in Fig.~\ref{fig:ocnet_pipeline}~(b).
	
	\vspace{.1cm}
	\noindent\textbf{Pyramid-OC}.
	We first apply a $3 \times 3$ convolution 
	to reduce the channels to $512$ in advance,
	then we feed the dimension reduced feature map to the Pyramid-OC and perform four different pyramid partitions ($1\times 1$ region,
	$2\times 2$ regions,
	$3\times 3$ regions,
	and $6 \times 6$ regions) on the 
	input feature map,
	and we concatenate the four different output object context feature maps output
	by the four parallel OCPs.
	Each one of the four object context feature maps has $512$ channels.
	We apply a $1 \times 1$ convolution to increase the channel of the input feature map
	from $512$ to $2048$ and concatenate it with all four object context feature maps.
	Lastly, we use a $1 \times 1$ convolution on the concatenated feature map with $4096$ channels and produce the final feature map with $512$ channels
	, which is not included in Fig.~\ref{fig:ocnet_pipeline}~(c).
	
	\vspace{.1cm}
	\noindent\textbf{ASP-OC}.
	We only perform the dimension reduction within the object context pooling branch, where we use a $3 \times 3$ convolution to reduce the channel to $256$. The output feature map from object context pooling module has $256$ channels.
	For the other four branches, we exactly follow the original ASPP module and apply a $1 \times 1$ convolution within the second above branch and $3 \times 3$ dilated convolution with different dilation rates ($12$, $24$, $36$) in the {\color{black}{three remaining}} parallel branches.
	We set the output channel as $256$ in all these four branches following the original settings~\citep{chen2017rethinking}.
	Lastly, we concatenate these five parallel output feature maps and use a $1 \times 1$ convolution
	to decrease the channel of the concatenated feature map from $1280$ to $256$,
	which is not included in Fig.~\ref{fig:ocnet_pipeline}~(d).

	\vspace{.1cm}
	\noindent\textbf{Discussion}.
	The concept of object context is also discussed in our another work: object contextual representations (OCR)~\citep{yuan2019object}.
	The main difference is that this work is focused on efficiently
	modeling the dense relations between \emph{pixel and pixel} while OCR mainly exploits the coarse segmentation maps
	to construct a set of object region representations and 
	models the dense relations between \emph{pixel and object regions}.

	\vspace{1cm}
	\section{Experimental Results}
\begin{figure*}[h!]
		\centering
		\begin{subfigure}[b]{0.5\textwidth}
			\includegraphics[width=\textwidth, height=4cm]{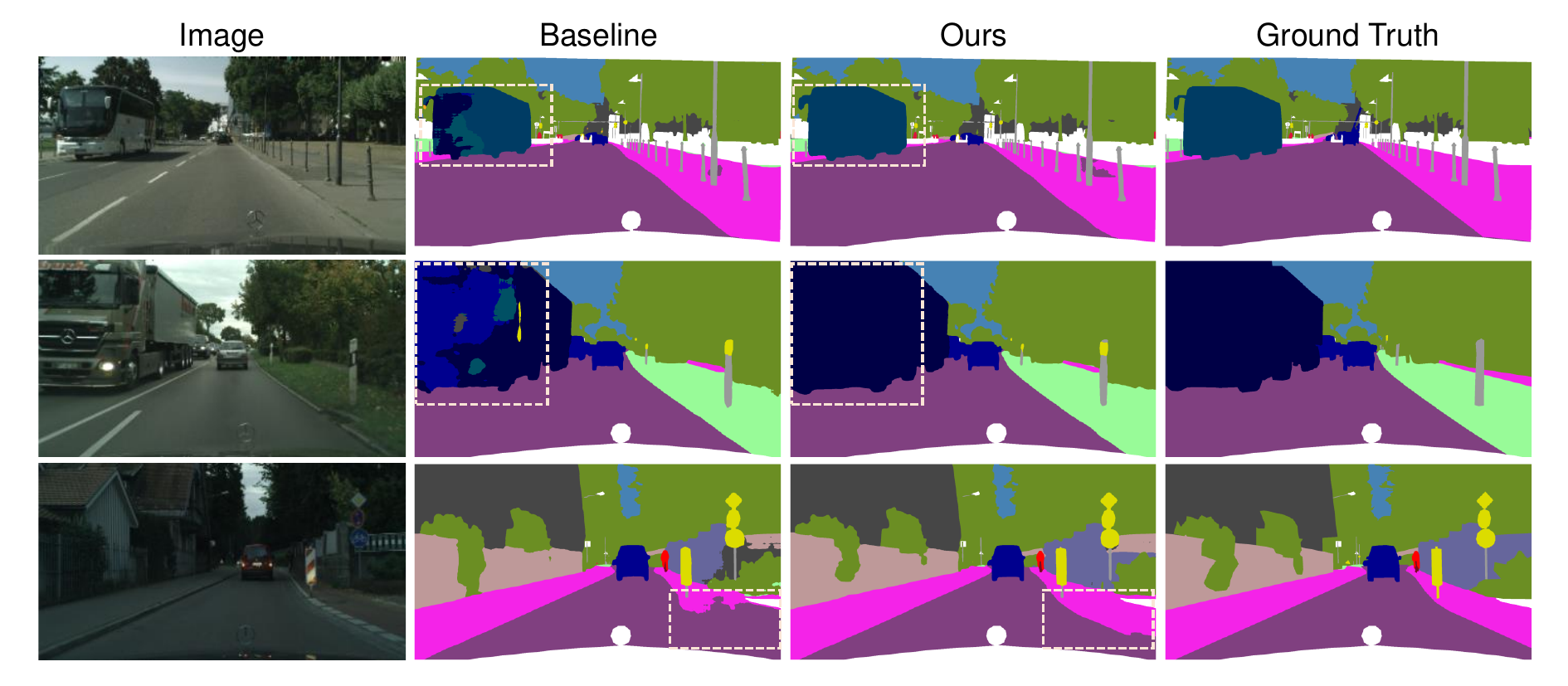}
			\caption{Cityscapes}
			\vspace*{3mm}
		\end{subfigure}
		\hspace{-0.5cm}
		\begin{subfigure}[b]{0.5\textwidth}
			\includegraphics[width=\textwidth, height=4cm]{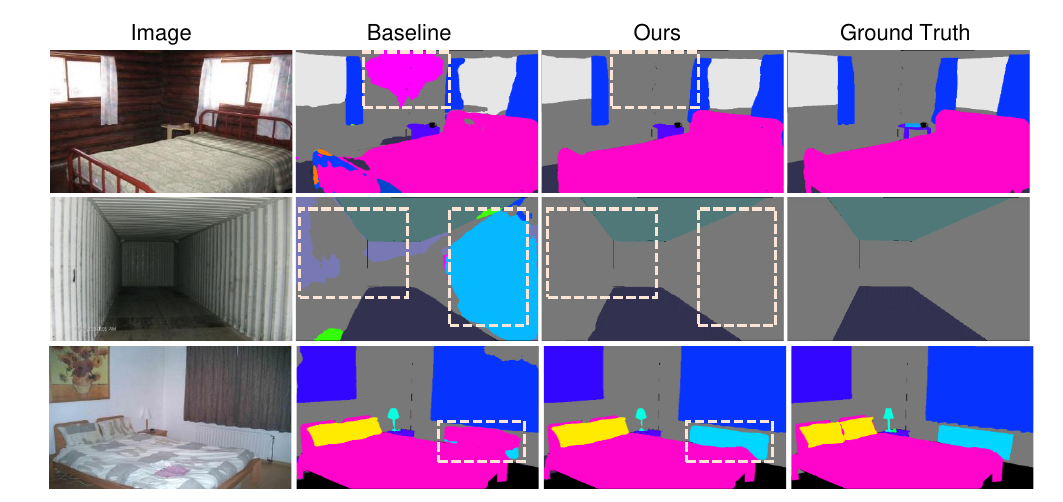}
			\caption{ADE20K}
			\vspace*{3mm}
		\end{subfigure}
		\begin{subfigure}[b]{0.5\textwidth}
			\includegraphics[width=\textwidth, height=4cm]{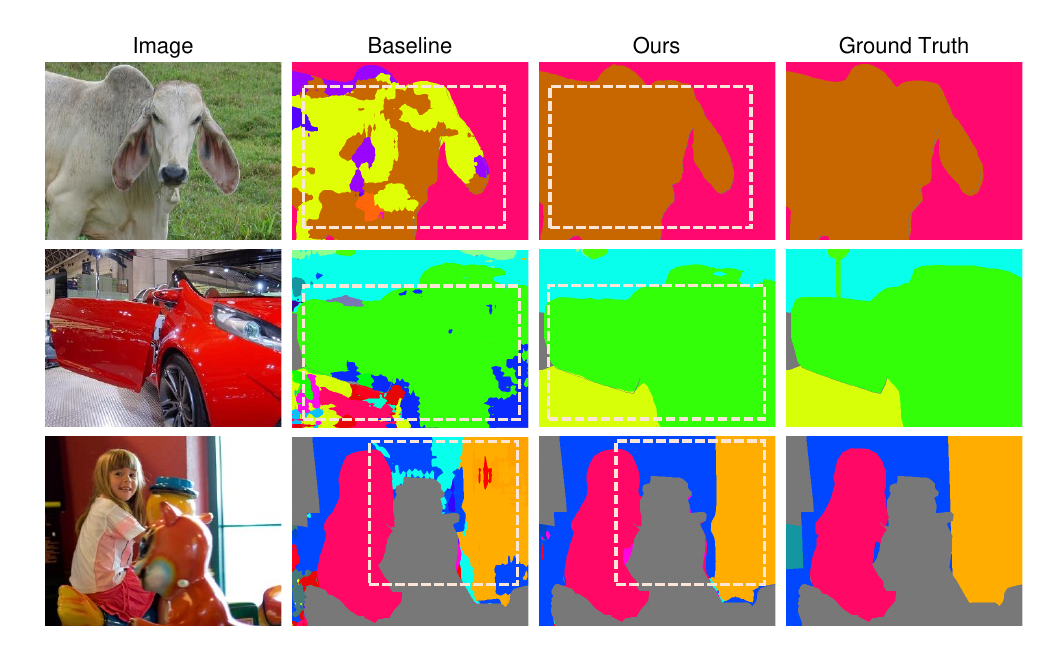}
			\caption{PASCAL-Context}
			\vspace*{3mm}
		\end{subfigure}
		\hspace{-0.5cm}
		\begin{subfigure}[b]{0.5\textwidth}
			\includegraphics[width=\textwidth, height=4cm]{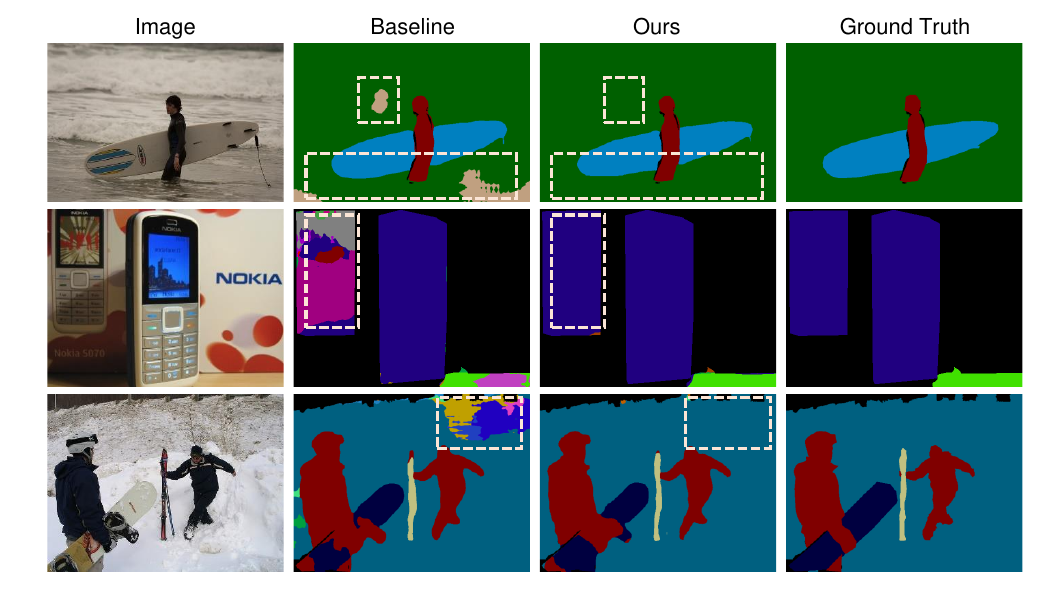}
			\caption{COCO-Stuff}
			\vspace*{3mm}
		\end{subfigure}
		\begin{subfigure}[b]{1\textwidth}
			\includegraphics[width=0.95\textwidth, height=9cm]{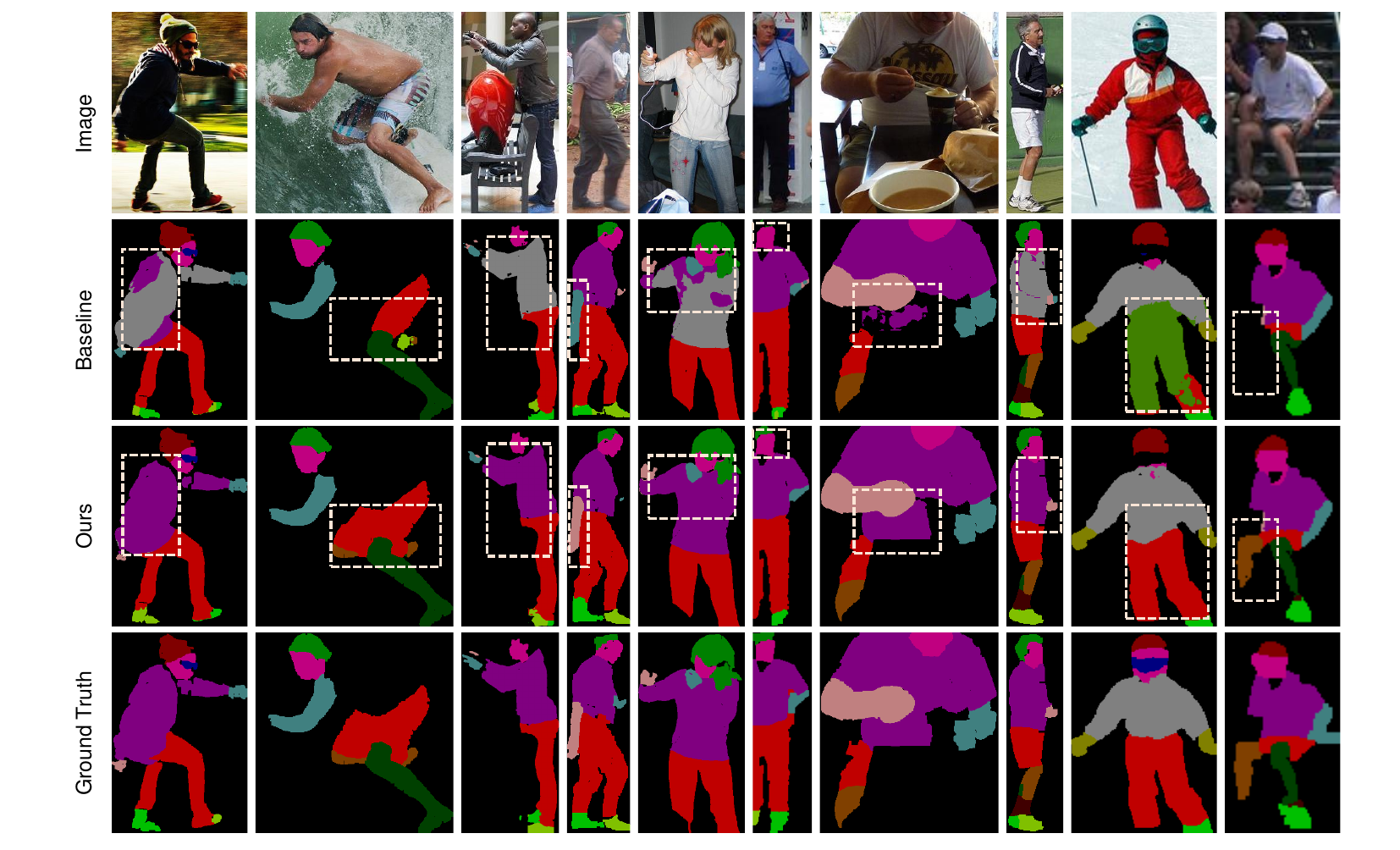}
			\caption{LIP}
		\end{subfigure}
		\caption{
			\small{
				Qualitative comparison on Cityscapes \texttt{val},
				ADE$20$K \texttt{val}, PASCAL-Context \texttt{test},
				COCO-Stuff \texttt{test} and LIP \texttt{val}.
				We choose dilated ResNet-$101$ as the baseline and further apply the Base-OC (ISA) on dilated ResNet-$101$ as our approach.
		}}
		\label{fig:seg-improve-1}
	\end{figure*}

	We evaluate our approach on five challenging
	semantic segmentation benchmarks.
	First, we study various components within our approach
	and compare our approach to some closely related mechanisms (Sec~\ref{ab_study}).
	Second, we compare our approach to the recent state-of-the-art 
	methods to verify that we achieve competitive performance (Sec~\ref{sota_study}).
	Last, we apply our approach on the conventional Mask-RCNN to 
	verify that our method generalizes well (Sec~\ref{detection_study}).
	Besides, we also illustrate the quantitative improvements along the boundary (Table~\ref{table:boundary_fscore}) and the qualitative improvements on various benchmarks (Fig.~\ref{fig:seg-improve-1}) based on our approach. 
	\textbf{
		\subsection{Datasets}
	}
	
	\noindent\textbf{Cityscapes}\footnote{\small{\url{https://www.cityscapes-dataset.com/}}}. 
	Cityscapes~\citep{cordts2016cityscapes} 
	contains $5,000$ finely annotated images with $19$ semantic classes.
	The images are in $2048\times 1024$ resolution
	and captured from 50 different cities.
	The training, validation, and test sets consist of 
	$2,975$,\; $500$,\; $1,525$ images, respectively.
	
	\vspace{.1cm}
	\noindent\textbf{ADE20K}\footnote{\url{https://groups.csail.mit.edu/vision/datasets/ADE20K/}}.
	ADE20K~\citep{zhou2017scene}
	is very challenging and it
	contains $22$K densely annotated images
	with $150$ fine-grained semantic concepts.
	The training and validation sets consist of
	$20$K, $2$K images, respectively.
	
	\vspace{.1cm}
	\noindent\textbf{LIP}\footnote{\url{http://sysu-hcp.net/lip/}}.
	LIP~\citep{Gong_2017_CVPR} is a large-scale dataset
	that focuses on semantic understanding of human bodies.
	It contains $50$K images with $19$ semantic human part labels
	and $1$ background label for human parsing.
	The training, validation, and test sets consist of 
	$30$K, $10$K, $10$K images, respectively.
	
	\vspace{.1cm}
	\noindent\textbf{PASCAL-Context}\footnote{\url{https://cs.stanford.edu/~roozbeh/pascal-context/}}.
	PASCAL-Context~\citep{mottaghi2014role} is a challenging
	scene parsing dataset that contains $59$ semantic
	classes and $1$ background class.
	The training set and test set consist of
	$4,998$ and $5,105$ images, respectively.
	
	\vspace{.1cm}
	\noindent\textbf{COCO-Stuff}\footnote{\url{https://github.com/nightrome/cocostuff}}.
	COCO-Stuff~\citep{caesar2018coco} is a challenging
	scene parsing dataset that contains $171$ semantic
	classes.
	The training set and test set consist of
	$9$K and $1$K images, respectively.
	
	\textbf{
		\subsection{Implementation Details}
	}
	
	\noindent\textbf{Training setting.}
	We initialize the parameters within
	the object context pooling module and the classification head randomly.
	We perform the polynomial learning rate policy with factor $(1-(\frac{iter}{iter_{max}})^{0.9})$.
	We set the weight on the final loss as $1$ and the weight on the auxiliary loss as $0.4$ following PSPNet~\citep{zhao2017pyramid}.
	{
	\color{black}{
	The auxiliary loss is applied on the representation output from stage-$3$ of ResNet-101 or
	the final representation of HRNetV2-W48. 
	}
	}
	We all use the \bnInplaceSync~\citep{Bulo_2018_CVPR} to synchronize the mean and standard-deviation of batch normalization across multiple GPUs.
	For the data augmentation, we perform random flipping horizontally, random scaling in the range of $[0.5, 2]$ and random brightness jittering within the range of $[-10, 10]$.
	More details are illustrated as following.
	
	\vspace{.1cm}
	For the experiments on \emph{Cityscapes}: 
	we set the initial learning rate as $0.01$, weight decay as $0.0005$, 
	crop size as $512\times 1024$ and batch size as $8$.
	For the experiments evaluated on \texttt{val}/\texttt{test},
	we set training iterations as $60$K/$100$K on \texttt{train}/\texttt{train}+\texttt{val} respectively.
	
	\vspace{.1cm}
	For the experiments on \emph{ADE20K}: 
	we set the initial learning rate as $0.02$, weight decay as $0.0001$, crop size as $520\times 520$, batch size as $16$ and training iterations as $150$K if not specified.
	
	\vspace{.1cm}
	For the experiments on \emph{LIP}: 
	we set the initial learning rate as $0.007$, weight decay as $0.0005$, crop size as $473\times 473$, batch size as $32$ and training iterations as $100$K if not specified.
	
	\vspace{.1cm}
	For the experiments on \emph{PASCAL-Context}: 
	we set the initial learning rate as $0.001$, weight decay as $0.0001$, crop size as $520\times 520$, batch size as $16$ and training iterations as $30$K if not specified.
	
	\vspace{.1cm}
	For the experiments on \emph{COCO-Stuff}: 
	we set the initial learning rate as $0.001$, weight decay as $0.0001$, crop size as $520\times 520$, batch size as $16$ and training iterations as $60$K if not specified.

	\textbf{
		\subsection{Ablation Study}
		\label{ab_study}
	}
	We choose the dilated ResNet-$101$
	as our backbone to conduct all
	ablation experiments,
	and we also use ResNet-$101$ alternatively
	for convenience.
	We choose the OCNet with Base-OC (ISA)
	as our default setting if not specified.
	
	\renewcommand{\arraystretch}{1.1}
	\begin{table}[t]
		\centering
		\small
		\caption{\small{Influence of ${P}_h$ and ${P}_w$,
				the order of global relation and local relation
				within the interlaced sparse self-attention on Cityscapes \texttt{val}.}}
		\resizebox{\linewidth}{!}
		{
			\begin{tabular}{l|c|c|c|c}
				\shline
				Method & ${P}_h$ & ${P}_w$ & Pixel Acc ($\%$) & mIoU ($\%$)  \\
				\shline
				Dilated ResNet-$101$    & - & -  & $96.08$ & $75.90$ \\
				\hline
				\multirow{7}{*}{+ Base-OC (ISA, Global+Local)} &  $4$ & $4$      & $96.30$ & $78.97$ \\ 
				& $4$ & $8$      & $96.31$ & $78.95$ \\ 
				& $8$ & $4$      & $96.32$ & $79.31$ \\ 
				& $8$ & $8$      & $\bf{96.33}$ & $\bf{79.49}$ \\ 
				& $8$ & $16$     & $96.29$ & $79.04$ \\ 
				& $16$ & $8$     & $96.19$ & $78.90$ \\ 
				& $16$ & $16$    & $96.32$ & $79.40$ \\ 
				\hline
				+ Base-OC (ISA, Local+Global) &  $8$ & $8$      & $96.26$ & $79.10$ \\
				\shline
			\end{tabular}
		}
		\label{table:ia_factor}
	\end{table}

	\renewcommand{\arraystretch}{1.2}
	\begin{table}[t]
		\centering
		\small
		\caption{\small{Comparison with
				multi-scale context scheme including
				PPM~\citep{zhao2017pyramid} and 
				ASPP~\citep{chen2017rethinking}
				on Cityscapes \texttt{val} and ADE20K \texttt{val}.
		}}
		\resizebox{\linewidth}{!}
		{
			\begin{tabular}{c|l|c|c} 
				\shline
				Dataset & Method      & Pixel Acc ($\%$)  & mIoU ($\%$)  \\
				\shline
				\multirow{5}{*}{Cityscapes}  & Dilated ResNet-$101$ & $96.08$ & $75.90$\\  
				& + PPM           &  $96.20$        & $78.50$ \\
				& + ASPP          &  $96.29$        & $79.10$ \\
				& + Base-OC (SA)        &  $96.32$    & $79.40$ \\
				& + Base-OC (ISA)         &  $\bf{96.33}$  & $\bf{79.49}$  \\ 
				\hline
				\multirow{5}{*}{ADE$20$K} &  Dilated ResNet-$50$ & $76.41$ & $34.35$ \\  
				& + PPM      & $80.17$            & $41.50$ \\ 
				& + ASPP     & $80.23$            & $42.00$ \\ 
				& + Base-OC (SA)            & $80.25$           & $42.05$  \\
				& + Base-OC (ISA)           & $\bf{80.27}$  & $\bf{42.11}$ \\
				\shline
			\end{tabular}
		}
		\label{table:compare-sa}
	\end{table}
	
	\renewcommand{\arraystretch}{1.2}
	\begin{table}[t]
		\centering
		\small
		\caption{\small{Comparison with
				SA with $2\times$ downsampling, RCCA and CGNL
				on Cityscapes \texttt{val}.}}
		\resizebox{\linewidth}{!}
		{
			\begin{tabular}{l|c|c|c|c|c}
				\shline
				Method &  SA-$2\times$ & RCCA  &  CGNL & Efficient Attention & Base-OC (ISA)  \\ 
				\shline
				mIoU & {\color{black}{$76.49_{\pm 0.35}$}} & $78.64_{\pm 0.12}$ & $77.06_{\pm 0.04}$ &  $78.25_{\pm 0.15}$ & $\bf{79.49_{\pm 0.11}}$ \\
				\shline
			\end{tabular}
		}
		\label{table:exp_ablation_down}
	\end{table}

	\renewcommand{\arraystretch}{1.2}
	\begin{table}[t]
		\centering
		\small
		\caption{\small{Efficiency comparison given input feature map of size  [$2048\times128\times128$]
				during inference stage.
				All results are based on Pytorch 0.4.1 with a single Tesla V100 with CUDA~$9.0$.}}
		\resizebox{\linewidth}{!}
		{
			\begin{tabular}{l|c|c|c}
				\shline
				Method              & Memory $\blacktriangle$   & GFLOPs $\blacktriangle$   &  Time  $\blacktriangle$  \\
				\shline
				PPM~\citep{zhao2017pyramid}      & $792$MB         & $619$     & $78$ms \\
				ASPP~\citep{chen2017rethinking}  & $\bf{331}$MB    & $492$     & $52$ms \\
				DANet~\citep{fu2018dual}         & $2339$MB        & $1110$    & $106$ms   \\
				RCCA~\citep{huang2018ccnet}      & $427$MB         & $804$     & $84$ms \\
				CGNL~\citep{yue2018cgnl}         & ${266}$MB    & $412$     & $43$ms  \\
				Efficient Attention~\citep{shen2018efficient}  & $\bf{214}$MB  & $\bf{331}$ & $\bf{35}$ms \\
				\hline
				Base-OC (SA)         & $2168$MB      & $619$       & $66$ms  \\
				Base-OC (ISA)        & ${301}$MB     & ${386}$  & ${42}$ms \\
				Pyramid-OC (SA)      & $2206$MB      & $880$       & $112$ms \\
				Pyramid-OC (ISA)     & $935$MB       & $595$       & $80$ms \\
				ASP-OC (SA)          & $348$MB       & $656$       & $67$ms  \\
				ASP-OC (ISA)         & $349$MB       & ${651}$     & ${65}$ms \\
				\shline
			\end{tabular}
		}
		\label{table:efficiency_compare}
	\end{table}
	
	\textbf{\renewcommand{\arraystretch}{1.2}
		\begin{table}[htb]
			\centering
			\footnotesize
			\caption{\small{Comparison to the pyramid extensions of object context pooling on Cityscapes \texttt{val}.}}
			\begin{tabular}{l|c} 
				\shline
				Method   &  mIoU ($\%$)  \\
				\shline
				Dilated ResNet-$101$    &  $75.90$    \\ \hline  
				+ Base-OC (SA)  &  $79.40$   \\ 
				+ Base-OC (ISA)  &  $79.49$  \\ 
				+ Pyramid-OC (SA)  &  $78.80$  \\
				+ Pyramid-OC (ISA)  &  $79.02$  \\ 
				+ ASP-OC (SA) &  $79.76$  \\
				+ ASP-OC (ISA)  &  $\bf{80.01}$  \\
				\shline
			\end{tabular}
			\label{table:pyramid_extension}
	\end{table}}
	
	\renewcommand{\arraystretch}{1.2}
	\begin{table*}[t]
		\centering
		\caption{\small{Category-wise improvements over the baseline based on Base-OC (ISA) in terms of boundary F-score on Cityscapes \texttt{val}.}}
		\resizebox{\linewidth}{!}
		{
			\begin{tabular}{c|l|ccccccccccccccccccc|c} 
				\shline
				thrs & method & \rotatebox{90}{road} & \rotatebox{90}{sidewalk} & \rotatebox{90}{building} & \rotatebox{90}{wall} & \rotatebox{90}{fence} &\rotatebox{90}{pole} & \rotatebox{90}{traffic light}&\rotatebox{90}{traffic sign}&\rotatebox{90}{vegetation}&\rotatebox{90}{terrian}&\rotatebox{90}{sky}&\rotatebox{90}{person}&\rotatebox{90}{rider}&\rotatebox{90}{car}&\rotatebox{90}{truck}&\rotatebox{90}{bus} & \rotatebox{90}{train} & \rotatebox{90}{motorcycle} & \rotatebox{90}{bicycle} & mean\\
				\shline
				\multirow{2}{*}{12px} & Dilated ResNet-$101$ & 92.6 & 80.7 & 87.8 & 57.1 & 57.3 & 83.6 & 77.2 & 82.3 & 91.0 & 62.3 & 90.2 & 79.7 & 81.7 & 92.2 & 78.5 & 89.8 & 90.4 & 82.9 & 80.3 & 80.9 \\
				& \textcolor{black}{+ Base-OC (ISA)} & \textbf{92.8} & 80.7 & \textbf{88.6} & \textbf{64.9} & \textbf{61.6} & \textbf{83.7} & 76.3 & 81.8 & 91.0 & \textbf{65.1} & \textbf{90.8} & 78.8 & 81.8 & \textbf{92.4} & \textbf{86.5} & \textbf{92.7} & \textbf{95.6} & 82.3 & 79.1 & \textbf{82.4} \\
				\hline
				\multirow{2}{*}{9px} & Dilated ResNet-$101$ & 91.7 & 78.9 & 85.7 & 56.0 & 55.9 & 82.4 & 76.1 & 81.0 & 88.9 & 61.0 & 89.3 & 78.1 & 80.3 & 91.0 & 78.0 & 89.4 & 90.2 & 82.4 & 78.1 & 79.7 \\
				& \textcolor{black}{+ Base-OC (ISA)} & \textbf{91.9} & 78.9 & \textbf{86.5} & \textbf{63.6} & \textbf{60.5} & 82.5 & 75.3 & 80.4 & 88.9 & \textbf{63.8} & 89.9 & 77.2 & 80.4 & 91.1 & \textbf{86.0} & \textbf{92.3} & \textbf{95.5} & 81.7 & 76.8 & \textbf{81.2} \\
				\hline
				\multirow{2}{*}{5px} & Dilated ResNet-$101$ & 88.9 & 73.6 & 79.5 & 53.1 & 52.7 & 78.9 & 72.1 & 76.7 & 82.2 & 57.4 & 85.9 & 73.1 & 75.9 & 86.9 & 77.0 & 88.5 & 90.0 & 80.9 & 71.7 & 76.1 \\
				& \textcolor{black}{+ Base-OC (ISA)} & 89.0 & 73.6 & \textbf{80.3} & \textbf{60.8} & \textbf{57.4} & 79.0 & 71.3 & 76.2 & 82.3 & \textbf{60.2} & \textbf{86.5} & 72.3 & 76.2 & 87.0 & \textbf{85.0} & \textbf{91.3} & \textbf{95.2} & 80.1 & 70.5 & \textbf{77.6} \\
				\hline
				\multirow{2}{*}{3px} & Dilated ResNet-$101$ & 84.3 & 65.8 & 70.7 & 50.0 & 49.6 & 72.1 & 65.7 & 69.1 & 72.1 & 53.4 & 79.1 & 64.9 & 70.3 & 79.9 & 75.8 & 87.1 & 89.6 & 79.0 & 63.6 & 70.7 \\
				& \textcolor{black}{+ Base-OC (ISA)} & 84.4 & 65.5 & \textbf{71.7} & \textbf{57.7} & \textbf{54.4} & 72.8 & 64.7 & 68.8 & 72.6 & \textbf{56.2} & 80.0 & 64.6 & 70.7 & 80.1 & \textbf{83.7} & \textbf{89.7} & \textbf{94.8} & 78.5 & 62.5 & \textbf{72.3} \\
				\shline
			\end{tabular}
		}
		\label{table:boundary_fscore}
	\end{table*}
	
	\vspace{.1cm}
	\noindent\textbf{Group numbers.}
	In order to study the influence of 
	group numbers within the Base-OC (ISA) scheme,
	we train the Base-OC (ISA) method by varying
	${P}_{h}$ and ${P}_{w}$, where we can determine the value of 
	${Q}_{h}$ and ${Q}_{w}$ according to $H={P}_{h}\times{Q}_{h}$
	and $W={P}_{w}\times{Q}_{w}$.
	In Table~\ref{table:ia_factor},
	we illustrate the results on 
	Cityscapes \texttt{val}.
	We can see that our approach with different group numbers consistently improves over the
	baseline and we get the best result with ${P}_{h}={P}_{w}=8$,
	therefore, we set ${P}_{h}={P}_{w}=8$ in all experiments
	by default setting if not specified.

	\vspace{.1cm}
	\noindent\textbf{Global+Local \emph{vs.} Local+Global.}
	We study the influence of the order of 
	global relation and local relation within Base-OC (ISA) module.
	We report the results in the $6^{\text{th}}$ row and $10^{\text{th}}$ row of Table~\ref{table:ia_factor}.
	We can see that both mechanisms
	improve over the baseline by a large margin.
	Applying the global relation first seems to be favorable.
	We apply the global relation first
	unless otherwise specified for all our experiments.
	Besides, we also compare the results with only sparse global
	attention or only local relation.
	We report the results (measured by mIoU):
	only global relation: $78.9\%$ and 
	only local relation: $77.2\%$,
	which verifies that the global relation is
	more important and the local relation is complementary
	with the global relation.

	\vspace{.1cm}
	\noindent\textbf{Comparison to multi-scale context.}
	We compare the proposed relational context scheme 
	to two conventional multi-scale context schemes including:
	PPM~\citep{zhao2017pyramid} and ASPP~\citep{chen2017rethinking}.
	
	We conduct the comparison experiments
	under the same training/testing settings,
	e.g., the same training iterations and batch size.
	We report the related results in Table~\ref{table:compare-sa}.
	Our reproduced PPM outperforms the original reported performance
	(Ours: $78.5\%$ vs. Paper~\citep{zhao2017pyramid}: $77.6\%$).
	Our approach 
	consistently outperforms both PPM and ASPP on 
	the evaluated benchmarks including Cityscapes and ADE$20$K.
	For example,
	Base-OC (ISA) outperforms the PPM
	by $0.99\%$/$0.61\%$ on Cityscapes/ADE$20$K, respectively 
	measured by mIoU.
	Compared to the ASPP,
	our approach is {\color{black}{more efficient}}
	according to the complexity comparison reported in Table~\ref{table:efficiency_compare}.
	
	Besides, we also compare the performance based on
	the conventional self-attention (SA) and the proposed
	interlaced sparse self-attention (ISA) in Table~\ref{table:compare-sa},
	and we can see that our ISA achieves comparable
	performance while being much more efficient.
	
	\vspace{.1cm}
	\noindent\textbf{Comparison to RCCA/CGNL/Efficient-Attention.}
	We compare our approach with several existing mechanisms
	that focus on addressing the efficiency 
	problem of self-attention/non-local,
	such as SA-$2\times$~\citep{wang2018non},
	RCCA~\citep{huang2018ccnet}, CGNL~\citep{yue2018cgnl} and Efficient Attention~\citep{shen2018efficient}.
	For SA-$2\times$, 
	we directly down-sample
	the feature map for $2\times$ before computing the dense 
	relation matrix.
	We evaluate all these mechanisms on the 
	Cityscapes \texttt{val} and report 
	the results in Table~\ref{table:exp_ablation_down}.
	We can see that our approach consistently
	outperforms all these three mechanisms, which verifies that 
	our approach is more reliable.
	We all report the average performance for fairness considering
	the mIoU variance of RCCA is large.
	
	\vspace{.1cm}
	\noindent\textbf{Complexity.}
	We compare the complexity of our approach with PPM~\citep{zhao2017pyramid},
	DANet~\citep{fu2018dual}, RCCA~\citep{huang2018ccnet}, CGNL~\citep{yue2018cgnl} and Efficient Attention~\citep{shen2018efficient} in this section.
	We report the GPU memory, GFLOPs and inference time when processing
	input feature map of size $2048\times128\times128$ under the same setting in Table~\ref{table:efficiency_compare}.
	We can see that 
	our approach based on ISA is much more efficient
	than most of the other approaches except the Efficient Attention scheme.
	{\color{black}{For example}}, the proposed Base-OC (ISA) is nearly $3\times$ faster 
	and saves more than $88\%$ GPU memory when compared with the DANet.
	Besides, our approach also requires less GPU memory/inference time
	than both RCCA and CGNL.

	\vspace{.1cm}
	\noindent\textbf{Pyramid extensions.}
	We study the performance with the two pyramid extensions
	including the Pyramid-OC and ASP-OC.
	We choose the dilated ResNet-$101$ as our baseline
	and summarize all related results
	in Table~\ref{table:pyramid_extension}.
	We can find that the ASP-OC consistently
	improves the performance for {\color{black}{both the SA scheme}} and ISA scheme
	while the Pyramid-OC slightly degrades the performance
	compared to the Base-OC mechanism.
	Accordingly,
	we only report the performance 
	with Base-OC (ISA) module and ASP-OC (ISA) module for 
	the following experiments
	if not specified.

	\renewcommand{\arraystretch}{1.2}
	\begin{table}[t]
		\centering
		\small
		\caption{\small{Comparison with state-of-the-arts on Cityscapes \texttt{test}.}}
		\resizebox{\linewidth}{!}
		{
			\begin{tabular}{l|c|c|c}
				\shline
				Method & Backbone & Validation & mIoU ($\%$)  \\
				\shline
				PSPNet~\citep{zhao2017pyramid}       & ResNet-$101$ & \xmark & $78.4$ \\ 
				PSANet~\citep{psanet}                & ResNet-$101$ & \xmark & $78.6$ \\ 
				AAF~\citep{aaf2018}                  & ResNet-$101$ & \xmark & ${79.1}$ \\ 
				RefineNet~\citep{lin2017refinenet}   & ResNet-$101$ & \cmark & $73.6$ \\ 
				DUC-HDC~\citep{wang2017understanding}& ResNet-$101$ & \cmark & $77.6$ \\ 
				DSSPN~\citep{Liang_2018_CVPR}        & ResNet-$101$ & \cmark & $77.8$ \\ 
				SAC~\citep{Zhang_2017_ICCV}          & ResNet-$101$ & \cmark & $78.1$ \\
				DepthSeg~\citep{Kong_2018_CVPR}      & ResNet-$101$ & \cmark & $78.2$ \\
				BiSeNet~\citep{yu2018bisenet}        & ResNet-$101$ & \cmark & $78.9$ \\
				DFN~\citep{Yu_2018_CVPR}             & ResNet-$101$ & \cmark & $79.3$ \\ 
				TKCN~\citep{wu2018treestructured}    & ResNet-$101$ & \cmark & $79.5$ \\
				PSANet~\citep{psanet}                & ResNet-$101$ & \cmark & $80.1$ \\
				DenseASPP~\citep{Yang_2018_CVPR}     & DenseNet-$161$ & \cmark & $80.6$\\
				CFNet~\citep{zhang2019co}             & ResNet-$101$ & \cmark & $79.6$ \\
				SVCNet~\citep{ding2019semantic}      & ResNet-$101$ & \cmark & $81.0$ \\
				DANet~\citep{fu2018dual}             & ResNet-$101$ & \cmark & $81.5$ \\
				BFP~\citep{ding2019bfp}              & ResNet-$101$ & \cmark & $81.4$ \\
				CCNet~\citep{huang2018ccnet}         & ResNet-$101$ & \cmark & $81.4$ \\
				Asymmetric NL~\citep{zhu2019asymmetric}  & ResNet-$101$ & \cmark & $81.3$ \\
				ACFNet~\citep{zhang2019acfnet}       & ResNet-$101$ & \cmark & $81.8$ \\
				SeENet~\citep{pang2019seenet}        & ResNet-$101$ & \cmark & $81.2$ \\
				SPGNet~\citep{cheng2019spgnet}        & $2\times$ ResNet-$50$ & \cmark & $81.1$ \\
				ACNet~\citep{fu2019acnet}            & ResNet-$101$ & \cmark & $\underline{82.3}$ \\
				HRNet~\citep{SunXLW19}               & HRNetV2-$48$ & \cmark & $81.6$ \\
				\hline
				OCNet (w/ Base-OC)                   & ResNet-$101$    & \cmark & $81.4$ \\
				OCNet (w/ ASP-OC)                    & ResNet-$101$    & \cmark & $81.9$ \\
				OCNet (w/ Base-OC)                   & HRNetV2-$48$    & \cmark & $82.0$ \\
				OCNet (w/ ASP-OC)                    & HRNetV2-$48$    & \cmark & $\bf{82.5}$ \\
				\shline 
			\end{tabular}
		}
		\label{table:exp_cs_sota}
	\end{table}
	
	\textbf{
		\subsection{Comparison to State-of-the-arts}
		\label{sota_study}
	}
	We choose the object context pooling module based on ISA, e.g., Base-OC (ISA) and ASP-OC (ISA), by default.
	We evaluate the performance of OCNet (w/ Base-OC) 
	and OCNet (w/ ASP-OC) on $5$ benchmarks
	and illustrate the related results as follows.
	
	\vspace{.1cm}
	\noindent\textbf{Cityscapes.}
	We report the comparison to the
	state-of-the-art methods on Cityscapes \texttt{test} 
	in Table~\ref{table:exp_cs_sota}, 
	and we apply OHEM, the multi-scale testing and flip testing 
	following the previous work.
	We apply our approach on both ResNet-$101$
	and HRNetV2-W$48$,
	and our approach
	achieves competitive performance
	with both backbones.
	For example, 
	we improve the performance of
	HRNetV2-W$48$ from $81.6\%$ to $82.5\%$
	with the ASP-OC module, which also
	outperforms the recent ACNet~\citep{fu2019acnet}.
	
	\renewcommand{\arraystretch}{1.2}
	\begin{table}[t]
		\centering
		\small
		\caption{\small{Comparison with state-of-the-arts on ADE20K \texttt{val}.}}
		\begin{tabular}{l|c|c}
			\shline
			Method  & Backbone & mIoU ($\%$)  \\
			\shline
			RefineNet~\citep{lin2017refinenet}   & ResNet-$101$  &  $40.20$ \\ 
			RefineNet~\citep{lin2017refinenet}   & ResNet-$152$  & $40.70$\\ 
			UperNet~\citep{xiao2018unified}      & ResNet-$101$  &  $42.66$ \\
			PSPNet~\citep{zhao2017pyramid}       & ResNet-$101$  &  $43.29$ \\ 
			PSPNet~\citep{zhao2017pyramid}       & ResNet-$152$  &  $43.51$ \\ 
			DSSPN~\citep{Liang_2018_CVPR}        & ResNet-$101$  &  $43.68$ \\ 
			PSANet~\citep{psanet}                & ResNet-$101$  &  $43.77$ \\ 
			SAC~\citep{Zhang_2017_ICCV}          & ResNet-$101$  &  $44.30$ \\
			SGR~\citep{NIPS2018_7456}            & ResNet-$101$  &  $44.32$ \\
			EncNet~\citep{Zhang_2018_CVPR}       & ResNet-$101$  &  $44.65$ \\
			GCU~\citep{li2018beyond}             & ResNet-$101$  &  $44.81$ \\
			CFNet~\citep{zhang2019co}            & ResNet-$101$  &  $44.89$ \\
			APCNet~\citep{he2019adaptive}        & ResNet-$101$  &  $45.38$ \\
			CCNet~\citep{huang2018ccnet}         & ResNet-$101$  &  $45.22$ \\
			Asymmetric NL~\citep{zhu2019asymmetric}  & ResNet-$101$  &  $45.24$ \\
			DANet~\citep{fu2018dual}             & ResNet-$101$  &  $45.22$ \\
			ACNet~\citep{fu2019acnet}            & ResNet-$101$  &  $\bf{45.90}$ \\
			\hline
			OCNet (w/ Base-OC)                   & ResNet-$101$     & $45.04$ \\
			OCNet (w/ ASP-OC)                    & ResNet-$101$     & $45.40$ \\
			OCNet (w/ Base-OC)                   & HRNetV2-$48$     & $45.10$ \\
			OCNet (w/ ASP-OC)                    & HRNetV2-$48$     & $45.50$ \\
			\shline
		\end{tabular}
		\vspace{-0.2cm}
		\label{table:ianet_sota_exp_ade20k}
	\end{table}
	
	\vspace{.1cm}
	\noindent\textbf{ADE20K.}
	In Table~\ref{table:ianet_sota_exp_ade20k},
	we compare our approach to the state-of-the-arts 
	on the ADE20K \texttt{val}.
	Our approach also achieves competitive performance,
	e.g., OCNet (w/ ASP-OC) based on ResNet-$101$
	and HRNetV2-W$48$ achieve $45.40\%$ and $45.50\%$ respectively,
	both are slightly worse than the recent ACNet~\citep{fu2019acnet}
	that exploits rich global context and local context.

	\vspace{.1cm}
	\noindent\textbf{PASCAL-Context.}
	As illustrated in Table~\ref{table:ianet_sota_exp_context},
	we compare our approach
	with the previous state-of-the-arts 
	on the PASCAL-Context \texttt{test}.
	We can find that our approach 
	significantly improves the performance of
	HRNet~\citep{SunXLW19} and achieves $56.2\%$
	with OCNet (w/ Base-OC), which 
	also outperforms most of the other previous approaches.
	
	\renewcommand{\arraystretch}{1.2}
	\begin{table}[t]
		\centering
		\small
		\caption{\small{Comparison with state-of-the-arts on PASCAL-Context \texttt{test}.}}
		\begin{tabular}{l|c|c}
			\shline
			Method  & Backbone & mIoU ($\%$)  \\
			\shline
			DeepLabv2~\citep{chen2018deeplab}    & ResNet-$101$  &  $45.7$\\
			UNet++~\citep{zhou2018unet++}        & ResNet-$101$  &  $47.7$\\
			PSPNet~\citep{zhao2017pyramid}       & ResNet-$101$  &  $47.8$\\
			CCL~\citep{ding2018ccl}              & ResNet-$101$  &  $51.6$\\
			EncNet~\citep{Zhang_2018_CVPR}       & ResNet-$101$  &  $51.7$\\
			SGR~\citep{NIPS2018_7456}            & ResNet-$101$  &  $52.5$\\
			DANet~\citep{fu2018dual}             & ResNet-$101$  &  $52.6$\\
			SVCNet~\citep{ding2019semantic}      & ResNet-$101$  &  $53.2$\\
			CFNet~\citep{zhang2019co}            & ResNet-$101$  &  $54.0$\\
			Dupsampling~\citep{tian2019dupsampling} & Xception-$71$  &  $52.5$\\
			HRNetV2~\citep{SunZJCXLMWLW19}       & HRNetV2-$48$     & $54.0$ \\
			APCNet~\citep{he2019adaptive}        & ResNet-$101$  & $\underline{54.7}$ \\
			EMANet~\citep{li2019ema}             & ResNet-$101$  &  $53.1$\\
			\hline
			OCNet (w/ Base-OC)                   & ResNet-$101$     & $54.1$ \\
			OCNet (w/ ASP-OC)                    & ResNet-$101$     & $54.0$ \\
			OCNet (w/ Base-OC)                   & HRNetV2-$48$     & $\bf{56.2}$ \\
			OCNet (w/ ASP-OC)                    & HRNetV2-$48$     & $56.0$ \\
			\shline
		\end{tabular}
		\label{table:ianet_sota_exp_context}
	\end{table}
	
	\vspace{.1cm}
	\noindent\textbf{LIP.}
	We compare our approach to the previous
	state-of-the-arts on LIP \texttt{val} and illustrate the 
	results in Table~\ref{table:ianet_sota_exp_lip}.
	The OCNet (w/ ASP-OC) based on HRNetV2-$48$
	achieves competitive performance $56.35\%$,
	which is slightly worse than the recent CNIF~\citep{wang2019cnif}.
	
	\renewcommand{\arraystretch}{1.2}
	\begin{table}[t]
		\centering
		\small
		\caption{\small{Comparison with state-of-the-arts on LIP \texttt{val}.}}
		\begin{tabular}{l|c|c}
			\shline
			Method  & Backbone & mIoU ($\%$)  \\
			\shline
			Attention+SSL~\citep{Gong_2017_CVPR} & ResNet-$101$  &  $44.73$\\ 
			MMAN~\citep{luo2018macro}            & ResNet-$101$  &  $46.81$ \\ 
			SS-NAN~\citep{Zhang_2017_ICCV}       & ResNet-$101$  &  $47.92$ \\ 
			MuLA~\citep{nie2018mutual}           & ResNet-$101$  &  $49.30$ \\ 
			JPPNet~\citep{liang2018look}         & ResNet-$101$  &  $51.37$ \\ 
			CE2P~\citep{liu2018devil}            & ResNet-$101$  &  $53.10$ \\
			HRNet~\citep{SunZJCXLMWLW19}         & HRNetV2-$48$  &  $55.90$ \\
			CNIF~\citep{wang2019cnif}            & ResNet-$101$  &  $\bf{56.93}$ \\
			\hline
			OCNet (w/ Base-OC)                   & ResNet-$101$     & $55.07$ \\
			OCNet (w/ ASP-OC)                    & ResNet-$101$     & $55.20$ \\
			OCNet (w/ Base-OC)                   & HRNetV2-$48$     & $56.20$ \\
			OCNet (w/ ASP-OC)                    & HRNetV2-$48$     & $56.35$ \\
			\shline
		\end{tabular}
		\label{table:ianet_sota_exp_lip}
	\end{table}

	\begin{figure*}[h!]
		\centering
		\begin{subfigure}[b]{0.91\textwidth}
			\includegraphics[width=\textwidth]{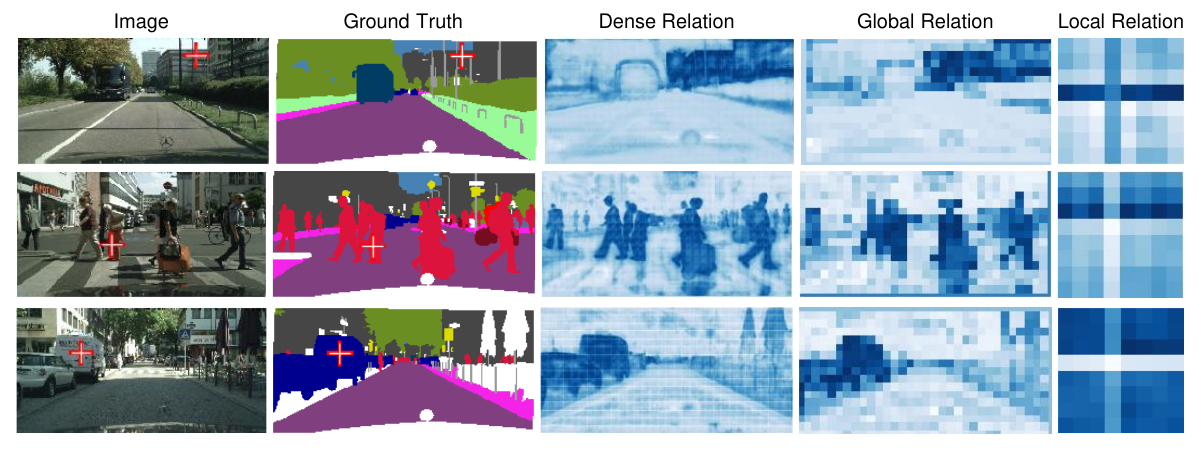}
			\caption{Cityscapes}
		\end{subfigure}
		\begin{subfigure}[b]{0.95\textwidth}
			\includegraphics[width=\textwidth]{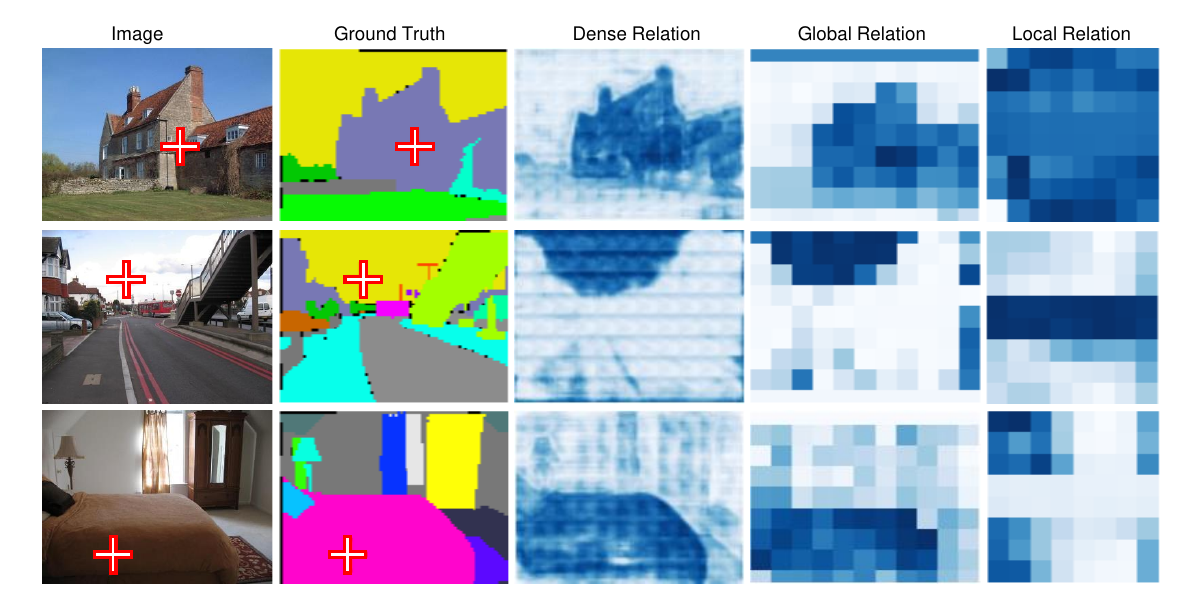}
			\caption{ADE20K}
		\end{subfigure}
		\begin{subfigure}[b]{0.95\textwidth}
			\includegraphics[width=\textwidth]{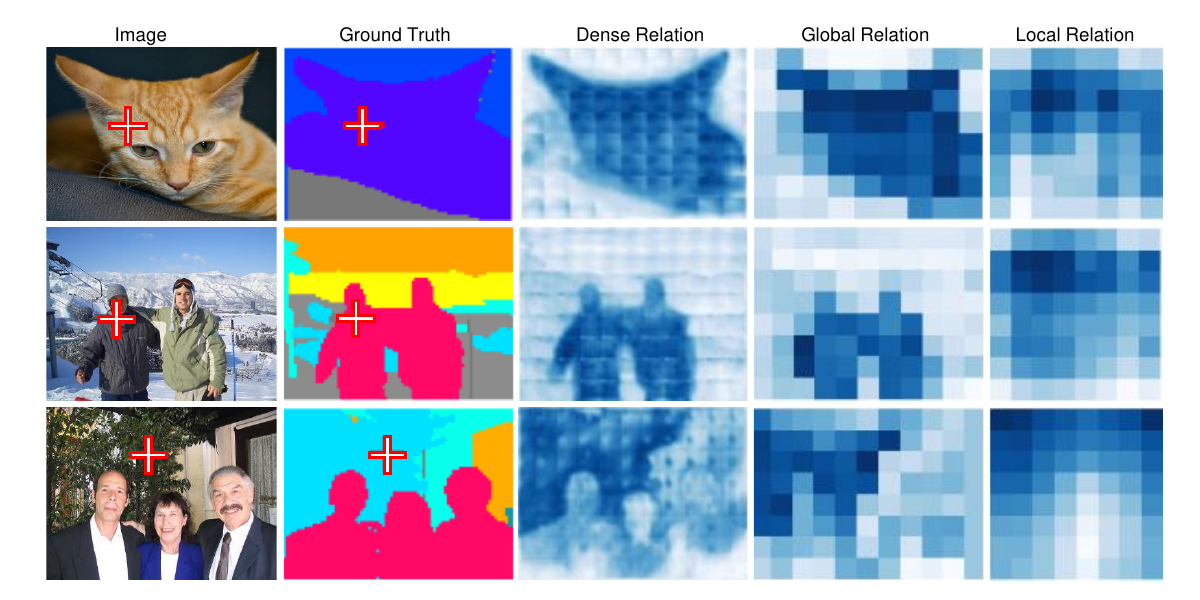}
			\caption{PASCAL-Context}
		\end{subfigure}
		\caption{\small{Visualization of the predicted dense relation matrices by OCNet on Cityscapes \texttt{val},
				ADE$20$K \texttt{val} and PASCAL-Context \texttt{test}.
				We apply the dilated ResNet-$101$ + Base-OC (ISA) to generate these relation matrices.
				We visualize both the global relation and the local relation for each selected pixel and we compute the dense relation as the product of the global relation and the local relation.
		}}
		\label{fig:dense_relation_1}
	\end{figure*}
	
	\begin{figure*}[h!]
		\centering
		\vspace{-2mm}
		\begin{subfigure}[b]{0.8\textwidth}
			\includegraphics[width=\textwidth]{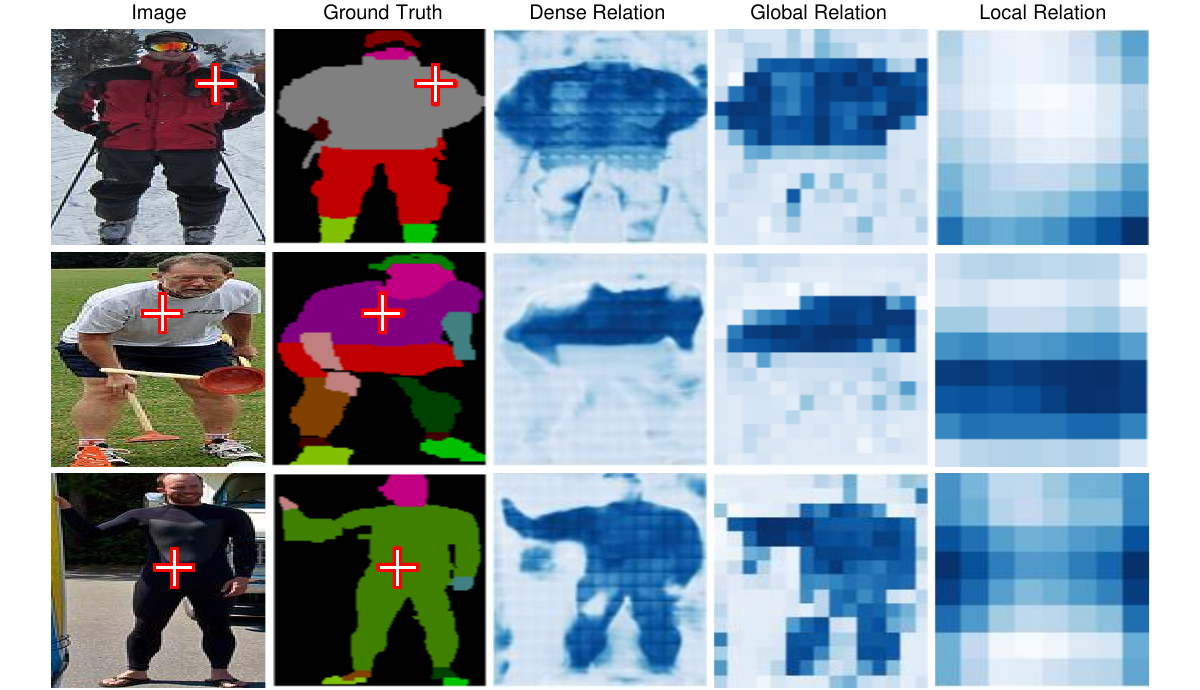}
			\caption{LIP}
		\end{subfigure}
		\begin{subfigure}[b]{0.8\textwidth}
			\includegraphics[width=\textwidth]{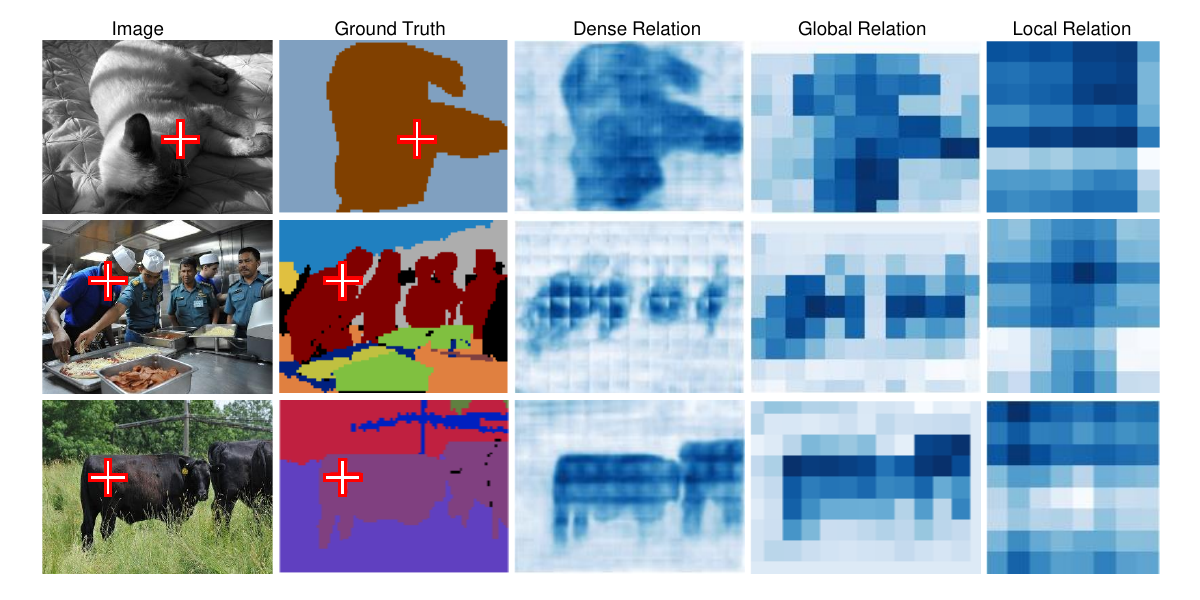}
			\caption{COCO-Stuff}
		\end{subfigure}
		\caption{\small{Visualization of the predicted dense relation matrices by OCNet on LIP \texttt{val} and COCO-Stuff \texttt{test}.
		We apply the dilated ResNet-$101$ + Base-OC (ISA) to generate these relation matrices.
		}}
		\label{fig:dense_relation_2}
	\end{figure*}
	
	\begin{figure*}[h!]
	\centering
		\includegraphics[width=0.9\textwidth]{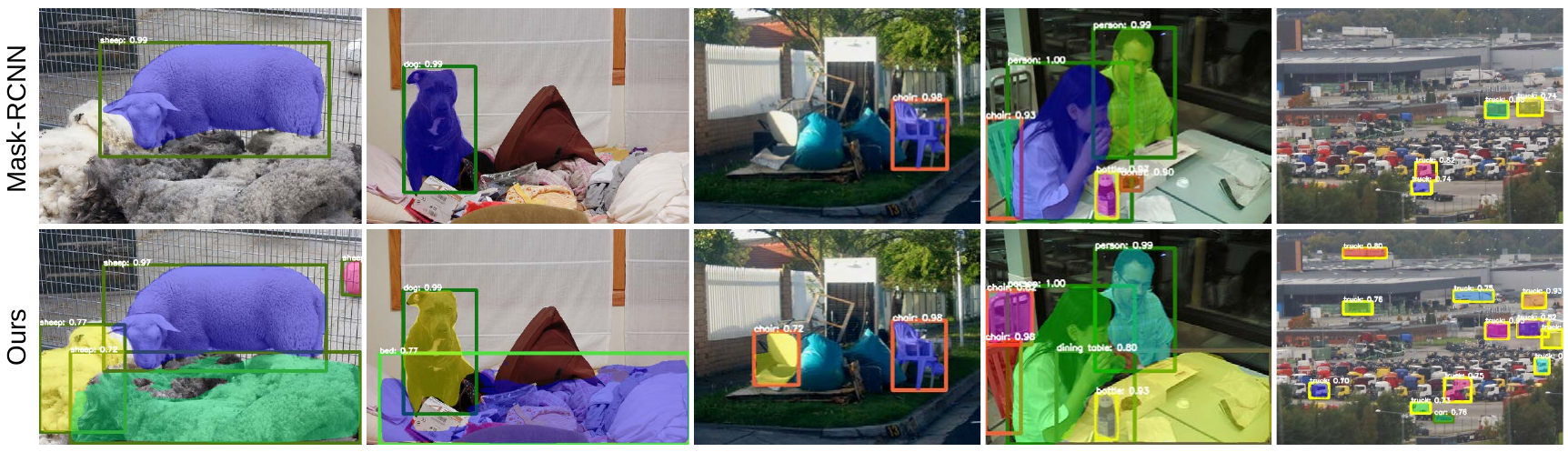}
		\caption{\small{Visualization of the object detection and instance segmentation results of Mask-RCNN~\citep{he2017mask} and our approach on the validation set of COCO (Best viewed in color).
		}}
		\label{fig:vis_coco}
	\end{figure*}

	\vspace{.1cm}
	\noindent\textbf{COCO-Stuff.}
	From Table~\ref{table:ianet_sota_exp_coco_suff},
	we can see that our method also achieves
	competitive performance $40.0\%$ on COCO-Stuff \texttt{test}, 
	which is comparable with the very recent 
	state-of-the-art method.

	\vspace{.1cm}
	\noindent\textbf{Visualization.}
	We visualize some examples of the global relation,
	local relation and dense relation predicted with our approach,
	e.g., OCNet based on dilated ResNet-$101$ + Base-OC (ISA),
	on different benchmarks in Fig.~\ref{fig:dense_relation_1} and Fig.~\ref{fig:dense_relation_2}.
	
	For all examples, 
	we down-sample the the ground-truth label map
	to match the size of the dense relation map, which is $\frac{1}{8}$
	of the input size.
	We choose the same group numbers ${P}_{h}={P}_{w}=8$ for all datasets, thus, the global relation matrix and the local relation matrix
	are of various shapes.
	For example, the dense relation matrix / global relation matrix / local relation matrix is of size $256\times128$ / $32\times16$ / $8\times8$ respectively for Cityscapes images.
	We generate the dense relation matrix by multiplying the local relation with 
	the global relation, and we can see that the estimated dense relation matrix
	puts its most relation weights on the pixels belonging to the 
	same category as the chosen pixel, which well approximates the 
	ground-truth object context.
	
	We compare the segmentation maps predicted with our approach
	and the baseline (dilated ResNet-$101$) to illustrate the qualitative improvements,
	and we visualize the results in Fig.~\ref{fig:seg-improve-1}.
	We can find that our
	method produces better segmentation maps compared with
	the baseline. We mark all of the improved regions with white
	dashed boxes.
	
	\vspace{.1cm}
	\noindent\textbf{Boundary Analysis.}
	We report the boundary improvements within 
	$3$, $5$, $9$ and $12$ pixels width based on our
	approach on Cityscapes \texttt{val} in Table~\ref{table:boundary_fscore},
	and we can find that our approach significantly
	improves the boundary quality for several object categories
	including wall, truck, bus, train and so on.

	\renewcommand{\arraystretch}{1.2}
	\begin{table}[t]
		\centering
		\small
		\caption{\small{Comparison with state-of-the-arts on COCO-Stuff \texttt{test}.}}
		\begin{tabular}{l|c|c}
			\shline
			Method  & Backbone & mIoU ($\%$)  \\
			\shline
			FCN~\citep{long2015fully}            & VGG-$16$      &  $22.7$\\
			DAG-RNN~\citep{shuai2017dag}         & VGG-$16$      &  $31.2$\\
			RefineNet~\citep{lin2017refinenet}   & ResNet-$101$  &  $33.6$\\
			CCL~\citep{ding2018ccl}              & ResNet-$101$  &  $35.7$\\
			SVCNet~\citep{ding2019semantic}      & ResNet-$101$  &  $39.6$\\
			DANet~\citep{fu2018dual}             & ResNet-$101$  &  $39.7$\\
			EMANet~\citep{li2019ema}             & ResNet-$101$  &  $39.9$\\
			ACNet~\citep{fu2019acnet}            & ResNet-$101$  &  $\bf{40.1}$\\
			\hline
			OCNet (w/ Base-OC)                   & ResNet-$101$     & $39.2$ \\ 
			OCNet (w/ ASP-OC)                    & ResNet-$101$     & $39.1$ \\ 
			OCNet (w/ Base-OC)                   & HRNetV2-$48$     & $40.0$ \\
			OCNet (w/ ASP-OC)                    & HRNetV2-$48$     & ${39.8}$ \\
			\shline
		\end{tabular}
		\label{table:ianet_sota_exp_coco_suff}
	\end{table}

	\textbf{
		\subsection{Application to Mask-RCNN}
		\label{detection_study}
	}
	
	\vspace{.1cm}
	\noindent\textbf{Dataset.}
	We use COCO~\citep{lin2014microsoft} dataset to evaluate our approach.
	The dataset is one of the most challenging datasets
	for object detection and instance segmentation,
	which contains $140$K images annotated
	with object bounding boxes and masks of 80 categories.
	We follow the COCO2017 split as in~\citep{he2017mask},
	where the training, validation and test sets contains
	$115$K, $5$K, $20$K images, respectively.
	We report the standard COCO metrics including Average Precision (AP),
	AP$_{50}$ and AP$_{75}$
	for both bounding boxes and masks.
	
	\vspace{.1cm}
	\noindent\textbf{Training settings.}
	We use Mask-RCNN~\citep{he2017mask} as baseline to conduct our experiments.
	Similar to~\citep{wang2018non}, we insert $1$ 
	non-local block or object context pooling module based on interlaced sparse self-attention
	before the last block of res-$4$ stage of
	the ResNet-$50$ FPN~\citep{lin2017fpn} backbone.
	All models are initialized with ImageNet pretrained weights
	and built upon open source toolbox~\citep{massa2018mrcnn}.
	We train the models using SGD with batch size of $16$
	and weight decay of $0.0001$.
	We conduct experiments using training schedules including
	``$1\times$ schedule'' and ``$2\times$ schedule''~\citep{massa2018mrcnn}.
	The $1\times$ schedule starts at a learning rate of $0.02$ and is decreased by a factor of $10$
	after $60$K and $80$K iterations and finally terminates at $90$K iterations.
	We train for $180$K iterations for $2\times$ schedule and 
	decreases the learning rate proportionally.
	The other training and inference strategies keep the same
	with the default settings in the~\citep{massa2018mrcnn}.
	
	\vspace{.1cm}
	\noindent\textbf{Results.}
	We report the results on COCO dataset in Table~\ref{table:exp_coco}.
	We can see that adding one non-local block~\citep{wang2018non} or interlaced sparse self-attention module 
	consistently improves the Mask-RCNN baseline by $\sim1$\%
	on all metrics involving both object detection and instance segmentation.
	Similar gains are observed for both $1\times$ schedule and $2\times$ schedule.
	For example, our approach improves the box AP/mask AP of Mask-RCNN
	from $38.7$/$34.9$ to $39.7$/$35.7$ with $2\times$ schedule.
	Especially, the performance of our approach is comparable
	with the non-local block on all metrics 
	while decreasing the computation complexity significantly.
	
	Last, we visualize the object detection and instance segmentation results of our approach and the Mask-RCNN on
	the validation set of COCO in Fig.~\ref{fig:vis_coco}. 
	We can find that
	our approach improves the Mask-RCNN consistently on all
	the examples. For example, the Mask-RCNN fails to detect multiple cars in the last example while our approach
	achieves better detection performance.
		
	\renewcommand{\arraystretch}{1.5}
	\begin{table}[t]
		\centering
		\small
		\caption{\small{Comparison with non-local~\citep{wang2018non} (NL)
				on the validation set of COCO.
				We use Mask-RCNN~\citep{he2017mask} as baseline
				and choose ResNet-$50$ FPN backbone for all models.}}
		\resizebox{\linewidth}{!}
		{
			\begin{tabular}{l|c|ccc|ccc}
				\shline
				Method          & Schedule & $\rm{AP}^{box}$ & $\rm{AP}^{box}_{50}$ & $\rm{AP}^{box}_{75}$ & $\rm{AP}^{mask}$ & $\rm{AP}^{mask}_{50}$  & $\rm{AP}^{mask}_{75}$\\
				\shline
				Mask-RCNN   & $1\times$ & $37.7$ & $59.2$ & $41.0$ & $34.2$ & $56.0$ & $36.2$ \\ 
				+ NL        & $1\times$ & $\bf{38.8}$ & $60.6$ & $42.3$ & $35.1$ & $\bf{57.4}$ & $37.3$ \\
				+ ISA      & $1\times$ & $\bf{38.8}$ & $\bf{60.7}$ & $\bf{42.5}$ & $\bf{35.2}$ & $57.3$ & $\bf{37.6}$\\
				\hline
				Mask-RCNN   & $2\times$ & $38.7$ & $59.9$ & $42.1$ & $34.9$ & $56.8$ & $37.0$ \\ 
				+ NL        & $2\times$ & $\bf{39.7}$ & $\bf{61.3}$ & $\bf{43.4}$ & $\bf{35.9}$ & $\bf{58.3}$ & $\bf{38.2}$ \\
				+ ISA      & $2\times$ & $\bf{39.7}$ & $61.1$ & $43.3$ & $35.7$ & $57.8$ & $38.1$ \\
				\shline
			\end{tabular}
		}
		\label{table:exp_coco}
	\end{table}

	\vspace{.5cm}
	\section{Conclusion}
	\textcolor{black}{
	In this paper, we present the object context
	that is capable of enhancing the object information via 
	exploiting the semantic relations between pixels.
	Our object context is more in line with the definition
	of the semantic segmentation that defines the category of each pixel
	as the category of the object that it belongs to.
	We propose two different kinds of implementations including:
	(i) dense relation based on the conventional self-attention scheme and
	(ii) sparse relation based on the proposed interlaced sparse self-attention scheme.
	We demonstrate that the effectiveness of our method
	on five challenging semantic segmentation benchmarks,
	e.g., Cityscapes, ADE20K, LIP,
	PASCAL-Context and COCO-Stuff.}
	We also extend our approach on Mask-RCNN to verify
	the advantage and we believe
	our approach might benefit various vision tasks
	through replacing the original self-attention or non-local scheme
	with our interlaced sparse self-attention mechanism.
	
	\vspace{.5cm}
	\section{Future work}
	\textcolor{black}{
    Although our object context scheme achieves competitive results on various benchmarks,
	there still exist many other important paths to construct richer context information.
	We illustrate three potential candidates:
	\begin{itemize}
	    \item use the co-occurring relations between different object categories to refine the coarse segmentation map,
        e.g., the ``rider'' tends to co-occur with the ``bicycle'', thus, we can refine the ``rider'' pixels being misclassified as ``person''.
        \item use the shape structure information to regularize the segmentation, e.g., the shape of ``bus'' tends to be quadrilateral, pentagon, or hexagon under various views, thus, we might use a set of prior shape masks to refine the predictions like the recent ShapeMask~\citep{Kuo_2019_ICCV}.
        \item the spatial location relation information, e.g., the ``keyboard'' is typically lying under the ``monitor'', thus, we can use the prior knowledge on the spatial relations between different objects to refine the predictions. Besides, there also exist some efforts~\citep{krishna2017visual} that focused on predicting the ``relationship'' between different objects from the input image directly. 
	\end{itemize}
	}
	
	{
	\clearpage
	\bibliographystyle{spbasic}\small
	\bibliography{ocnet}
	}

	\vspace{5mm}
	\section*{Appendix}

	\vspace{2mm}
	\subsection*{\textbf{A. More discussions on the benefits of \emph{object context}.}}

\begin{figure*}[t]
		\begin{centering}
		\setlength{\tabcolsep}{0em}
		\renewcommand{\arraystretch}{0}
		\par\end{centering}
		\begin{centering}
		\begin{tabular}{@{}c@{}c@{\hskip 0.05in}c@{}c@{}}
		
			\includegraphics[width=0.24\textwidth, height=0.09\textheight]{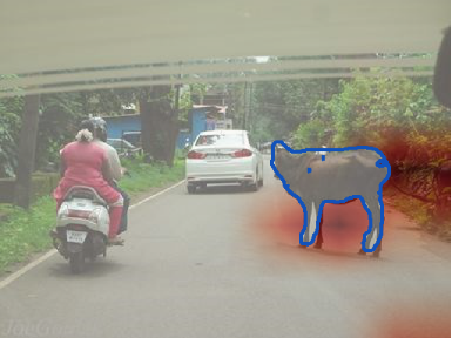} & {\footnotesize{}}
			\includegraphics[width=0.24\textwidth, height=0.09\textheight]{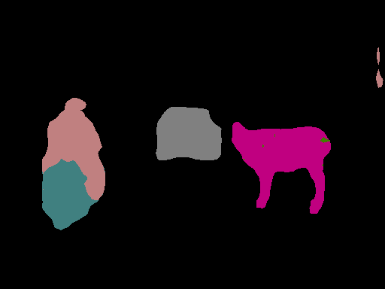}& {\footnotesize{}}
			\includegraphics[width=0.24\textwidth, height=0.09\textheight]{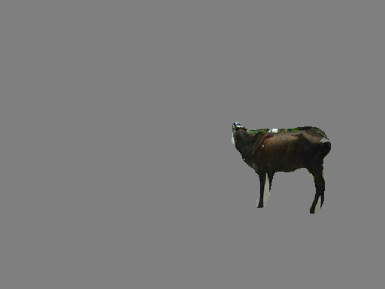} & {\footnotesize{}}
			\includegraphics[width=0.24\textwidth, height=0.09\textheight]{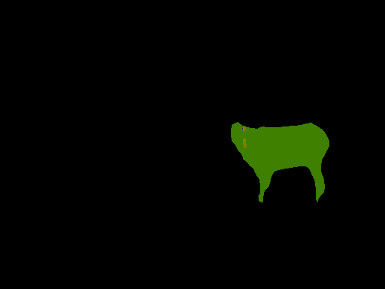} 
			\tabularnewline
			(a) Context explanation & (b) Semantic segm. & (c) Image w/o context & (d) Segm. w/o context
			
		\end{tabular}
		\par\end{centering}
		\caption{Effect of background context on semantic segmentation and the context explanation provided by grid saliency for an erroneous prediction, the image is taken from MS COCO~\cite{lin2014microsoft}. The grid saliency (a) shows the responsible context for misclassifying the cow (green) as horse (purple) in the semantic segmentation (b). It shows the training bias that horses are more likely on road than cows. Removing the background context context (c) yields a correctly classified cow (d).}
		\label{fig:context_cow}
	\end{figure*}

	We give a further explanation on why we believe OC is superior to the previous 
	two representation methods including PPM~\citep{zhao2017pyramid} and ASPP~\citep{chen2017rethinking} as following:

	\begin{itemize}
	\item In theory, enhancing the object information in the context can decrease the variance of the context information, in other words,
	the context of PPM and ASPP suffers from larger variance than the OC context. 
	Because the OC context only contains the variance of the \{object information\} while the context of PPM/ASPP further contains the variance of \{object information, useful background information, irrelevant background information \}. 
	The recent study~\citep{hoyer2019grid,shetty2019not} has verified that the overuse of the noisy context information based on PPM
	suffers from poor generalization ability.
	For example, the ``cow'' pixels might be mis-classified as ``horse'' pixels 
	when the ``cow'' appears on the road. 
	We directly use the Fig.~\ref{fig:context_cow} from~\citet{hoyer2019grid} to support our point.
	In summary, explicitly enhancing the object information might decrease the variance
	of the context information, thus, increases the generalization ability of model.
	\item In experiments, according to the results in the Table 2 (in the paper), we have verified that the OCNet outperforms both PSPNet and DeepLabv3 under the fair comparison settings.
	\end{itemize}

	\vspace{2mm}
	\subsection*{\textbf{B. Formulation of PPM context.}}
	We illustrate the definition of the context $\mathcal{I}_i$ based on PPM~\citep{zhao2017pyramid} scheme:
    \begin{ceqn}
    \begin{align}
    \mathcal{I}_i = \{j\in\mathcal{I}~|~\lfloor \frac{(j - 1)k}{N} \rfloor  = \lfloor \frac{(i - 1)k}{N} \rfloor \}, 
    \label{eqn:ppm_context}
    \end{align}
    \end{ceqn}
    where $k \in \{2, 3, 6\}$
	represents different pyramid region partitions.
	Such context is a aggregation of the pixels the the same quotient.
    
    \vspace{2mm}
    \subsection*{\textbf{C. Formulation of Permutation Matrix.}}
    We illustrate the definition of each value $p_{i,j}$ in the permutation matrix $\mathbf{P}$:
	\begin{ceqn}
    \begin{equation}
    p_{i,j} = 
    \begin{cases}
    1, & j = ((i-1) \bmod P) \times P + \lfloor\frac{i-1}{P}\rfloor + 1; \label{eq:0} \\
    0, & \mathrm{otherwise},
    \end{cases}
    \end{equation} 
    \end{ceqn}
    where, according to $\mathbf{W} = \mathbf{W}^l \mathbf{P}^{\top} \mathbf{W}^g \mathbf{P}$,
    we permute the $i$-th column of $\mathbf{W}^g$ / $\mathbf{W}^l$ to the $j$-th column if $p_{i,j}=1$ / $p_{j,i}=1$
    when multiplying permutation matrix $\mathbf{P}$ / $\mathbf{P}^{\top}$ on the right side of $\mathbf{W}^g$ / $\mathbf{W}^l$ respectively.
		
	\vspace{2mm}
	\subsection*{\textbf{D. Why the sparse relation is more efficient?}}
	For the convenience of analysis,
	we rewrite the mathematical formulation of computing the
	context representations (w/o considering the transform functions $\delta(\cdot)$ and $\rho(\cdot)$)
	based on dense relation scheme
	and sparse relation scheme as following.
	The formulation of dense relation scheme is $\mathbf{Z} = \mathbf{W} \mathbf{X}$
	and the formulation of sparse relation scheme is
	$\mathbf{Z} =(\mathbf{W}^{l}\mathbf{P}^{\top}\mathbf{W}^{g}\mathbf{P})\mathbf{X}$.
	We can see that the formulation of the sparse relation scheme
	still requires $\mathcal{O}(N^2)$ GPU memory to store the reconstructed
	dense relation matrix $\mathbf{W}^{l}\mathbf{P}^{\top}\mathbf{W}^{g}\mathbf{P}$.
	To avoid such expensive GPU memory consumption,
	we rewrite the formulation of the sparse relation scheme as 
	$\mathbf{Z} =\mathbf{W}^{l}(\mathbf{P}^{\top}(\mathbf{W}^{g}(\mathbf{P}\mathbf{X})))$
	according to the associative laws.
	Because both $\mathbf{W}^{l}$ and $\mathbf{W}^{g}$
	are sparse block matrices and each block is independent from the other blocks,
	we compute the multiple block matrices concurrently via transforming
	these block matrices to align on the batch dimension.
	Besides, we also implement the permutation matrix via the combination of permute
	and reshape operation provided in PyTorch.
	More details are illustrated in the discussion in Sec.~\ref{instantiations} and Algorithm~\ref{fig:code}

\vspace{2mm}
\subsection*{\textbf{E. Intuitive example of the sparse relation scheme.}}\label{dense_relation}

\begin{figure}[t]
\includegraphics[width=.5\textwidth]{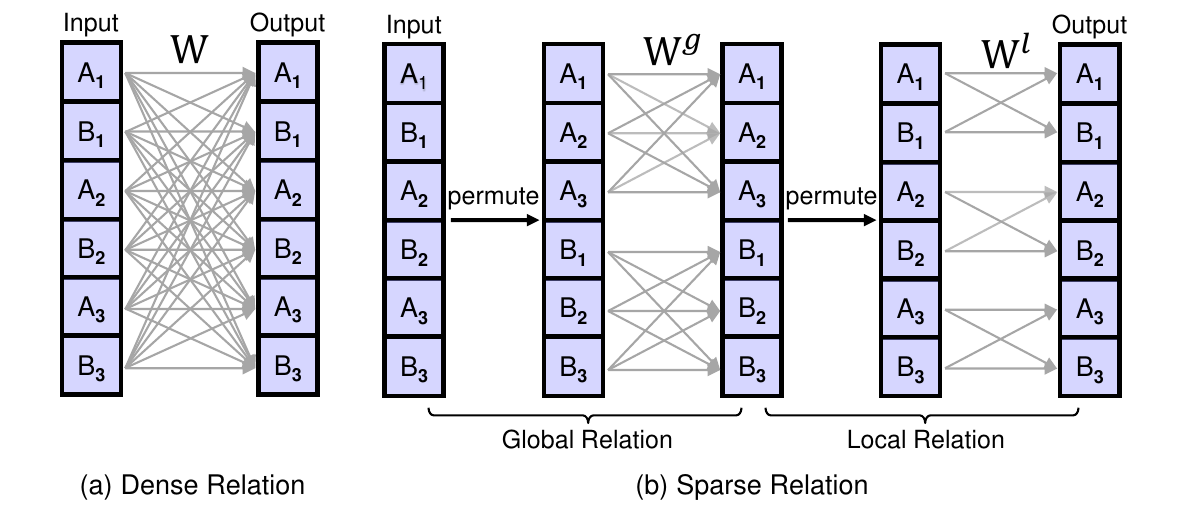}
\caption{\small{
\textbf{
Illustrating how the sparse relation approximates the dense relation.}
We use $\mathrm{A}_1,\mathrm{B}_1,...,\mathrm{B}_3$ to represent the different input positions.
The gray arrows represent the information propagation path 
from one input position to one output position.
In (a) Dense Relation, each output position connects with 
all input positions directly, thus, the relation matrix is fully dense.
We use the dense relation matrix $\mathbf{W}$ to record
the weights on all connections.
In (b) Sparse Relation, we have two relation matrices
and each relation matrix only contains the sparse connections to a small
set of selected pixels,
and the combination of the two sparse connections ensures that
each output position has direct or indirect relations with all input positions.
We use the two sparse relation matrices $\mathbf{W}^g$ and $\mathbf{W}^l$ 
to record the weights on all sparse connections.
}
}
\label{fig:connection}
\end{figure}

We use an one-dimensional example in Fig.~\ref{fig:connection}
to explain why the combination of two sparse relation 
matrices are capable to approximate the dense relation matrix.
In other words, both dense relation and sparse relation
ensure that each output position is connected with all input positions.
Specifically, in Fig.~\ref{fig:connection} (b), 
the output position $\rm{A}_1$ has direct relations with 
$\{\rm{A}_2, \rm{A}_3, \rm{B}_1\}$ and indirect relations
with $\{\rm{B}_2, \rm{B}_3\}$ via $\rm{B}_1$.

\vspace{2mm}
\subsection*{\textbf{F. Complexity Analysis.}}\label{complexity_proof}

	We illustrate the proof of the complexity of interlaced sparse self-attention scheme:
	\begin{proof}.
	The shapes of the input \& output are: $\mathbf{X}$ is of shape $HW\times C$, $\theta(\mathbf{X}), \phi(\mathbf{X}), \delta(\mathbf{X}) \in \mathbb{R}^{HW\times \frac{C}{2}}$, $\rho(\mathbf{X}) \in \mathbb{R}^{HW\times C}$.
	In the formulation of the ISA's global relation stage,
	the overall complexity of $\theta(\cdot)$, $\phi(\cdot)$, $\delta(\cdot)$, and $\rho(\cdot)$ are $\mathcal{O}(HWC^2)$.
	The overall complexity of $\theta(\mathbf{X}^{g}_p)\phi(\mathbf{X}^{g}_p)^{\top}$ and $\mathbf{W}_p^{g} \delta(\mathbf{X}^{g}_p)$ are $\mathcal{O}((\frac{HW}{P_{h} P_{w}})^{2}C)$ within each group of positions. There exist $P_{h}P_{w}$ groups in the global relation stage
	and $P_{h}P_{w}$.
	Thus, the overall complexity of the global relation stage of ISA is:
	\begin{ceqn}
	\begin{align}
	\label{eq:isa_global}
	\begin{split}
	T({\rm{ISA/global}}) = T(\theta(\cdot)) + T(\phi(\cdot)) + T(\delta(\cdot))  + T(\rho(\cdot)) \\
	+ P_{h}P_{w} T(\theta(\mathbf{X}^{g}_p)\phi(\mathbf{X}^{g}_p)^{\top}) 
	= \mathcal{O}(HWC^2+(HW)^2{C}\frac{1}{{P}_{h}{P}_{w}}),
	\end{split}
	\end{align}
	\end{ceqn}
	
	Similarly, we can get the complexity of the lobal relation stage in ISA:
	\begin{ceqn}
	\begin{align}
	\label{eq:isa_local}
	\begin{split}
	T({\rm{ISA/local}})
	= \mathcal{O}(HWC^2+(HW)^2{C}\frac{1}{{Q}_{h}{Q}_{w}}),
	\end{split}
	\end{align}
	\end{ceqn}
	
	In summary, we can compute the final complexity of ISA via adding $T({\rm{ISA/global}})$
	and $T({\rm{ISA/local}})$,
	\begin{ceqn}
	\begin{align}
	\label{eq:compl_isa}
	\begin{split}
	T({\rm{ISA}}) &= \mathcal{O}(HWC^2+(HW)^2{C}(\frac{1}{{P}_{h}{P}_{w}}+\frac{1}{{Q}_{h}{Q}_{w}})) \\
	\end{split}
	\end{align}
	\end{ceqn}
	where we can achieve the minimized computation complexity of $\mathcal{O}(HWC^2+(HW)^{\frac{3}{2}}{C})$ 
	when ${P_{h}P_{w}} = {Q_{h}Q_{w}}$ is satisfied (according to \emph{arithmetic mean} $\geq$ \emph{geometric mean}).
	\end{proof}

\vspace{2mm}
\subsection*{\textbf{G. Illustrating the Permutation Scheme of ISA.}}
\begin{figure*}[t]
	\includegraphics[width=\textwidth]{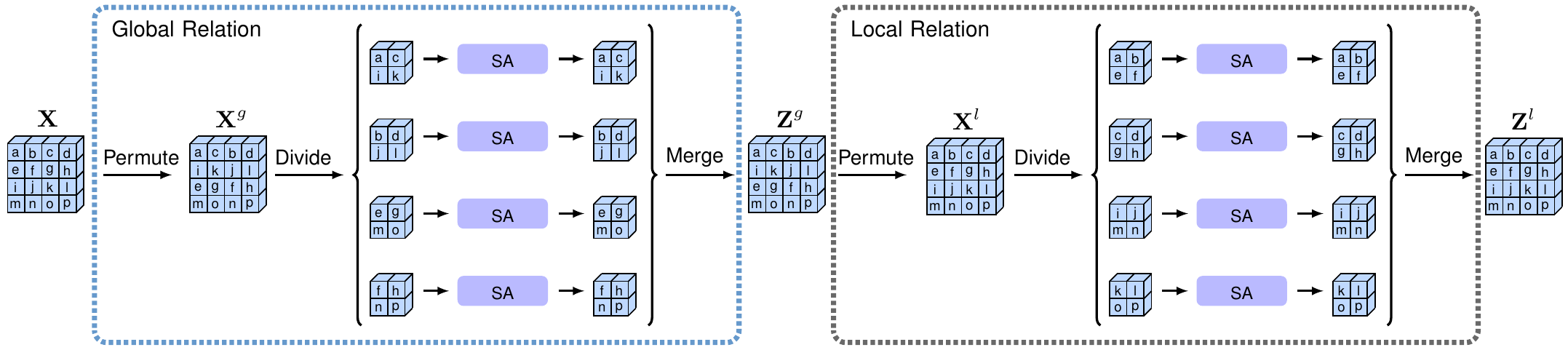}
	\caption{\small{
			\textbf{Illustrating the Interlaced Sparse Self-Attention with Indices Permutation}.
			We mark the positions in the input feature map with the indices from $a$ to $p$, e.g., the index $a$ represents the spatial position $(1,1)$ and the index $p$ represents the spatial position $(4, 4)$.
			We illustrate how we permute the indices in all stages as following:
			(i) For the \texttt{Permute} in the global relation stage, 
			we permute the positions according to the remainder of the indices divided by the group numbers, e.g., 2 for both height and width dimension.
			For example, all positions including $\{(1, 1), (1, 3), (3, 1), (2, 2)\}$ share the same remainder $(1, 1)$ when we divide the indices by 2 for both dimensions, thus, we group the positions $\{\rm{a}, \rm{c}, \rm{i}, \rm{k}\}$ together.
			Similarly, we get the other $3$ groups of positions: $\{\rm{b}, \rm{d}, \rm{j}, \rm{l}\}$ (share the same remainder $(1, 0)$), $\{\rm{e}, \rm{g}, \rm{m}, \rm{o}\}$ (share the same remainder $(0, 1)$) and $\{\rm{f}, \rm{h}, \rm{n}, \rm{p}\}$ (share the same remainder $(0, 0)$).
			(ii) For the \texttt{Permute} in the local relation stage,
			we permute the positions according to the quotient of the indices divided by the group numbers $(2, 2)$. Similarly, we get $4$ groups of positions:
			$\{\rm{a}, \rm{b}, \rm{e}, \rm{f}\}$ (share the same quotient $(0, 0)$), $\{\rm{c}, \rm{d}, \rm{g}, \rm{h}\}$ (share the same quotient $(0, 1)$), $\{\rm{i}, \rm{j}, \rm{m}, \rm{n}\}$ (share the same quotient $(1, 0)$) and $\{\rm{k}, \rm{l}, \rm{o}, \rm{p}\}$ (share the same quotient $(1, 1)$). 
	}}
\label{fig:isa_permute_index}
\end{figure*}

To help the readers to understand how we select and permute the indices within Interlaced Sparse Self-Attention,
we use an example in Fig.~\ref{fig:isa_permute_index} to explain the details.

\vspace{2mm}
\subsection*{\textbf{H. More details of Pyramid-OC.}}
We explain the details of Pyramid-OC as following:
Given an input feature map $\mathbf{X}$ of shape ${H\times W\times C}$, 
we first divide it into $k\times k$ groups ($k\in \{1,2,3,6\}$) following 
the pyramid partitions of PPM~\citep{zhao2017pyramid}:
\begin{ceqn}
\begin{align}
\mathbf{X}  \rightarrow 
\begin{bmatrix}
    \mathbf{X}_{1,1} & \mathbf{X}_{1,2} & \cdots & \mathbf{X}_{1,k} \\
    \mathbf{X}_{2,1} & \mathbf{X}_{2,2} & \cdots & \mathbf{X}_{2,k} \\
    \vdots     &\vdots      & \ddots & \vdots \\
    \mathbf{X}_{k,1} & \mathbf{X}_{k,2} & \cdots & \mathbf{X}_{k,k} \\
\end{bmatrix},
\end{align}
\end{ceqn}
where each $\mathbf{X}_{i,j}$ is of shape ${\frac{H}{k}\times \frac{W}{k}\times C}, \forall~i, j\in \{1,2,..,k\}$.
We apply the object context pooling (OCP) on each group $\mathbf{X}_{i,j}$ (note that the parameters of OCP are shared across the groups within the same partition) to compute the context representations, then, we concatenate the context representations to obtain the output feature $\mathbf{Z}^{k}$ of shape ${H\times W\times C}$:
\begin{ceqn}
\begin{align}
\begin{bmatrix}
    \mathrm{OCP}(\mathbf{X}_{1,1}) & \mathrm{OCP}(\mathbf{X}_{1,2}) & \cdots & \mathrm{OCP}(\mathbf{X}_{1,k}) \\
    \mathrm{OCP}(\mathbf{X}_{2,1}) & \mathrm{OCP}(\mathbf{X}_{2,2}) & \cdots & \mathrm{OCP}(\mathbf{X}_{2,k}) \\
    \vdots     &\vdots      & \ddots & \vdots \\
    \mathrm{OCP}(\mathbf{X}_{k,1}) & \mathrm{OCP}(\mathbf{X}_{k,2}) & \cdots & \mathrm{OCP}(\mathbf{X}_{k,k}) \\
\end{bmatrix}
  \rightarrow  \mathbf{Z}^{k}.
\end{align}
\end{ceqn}

We compute four different context feature maps $\{\mathbf{Z}^{1}, \mathbf{Z}^{2}, \mathbf{Z}^{3}, \mathbf{Z}^{6}\}$ based on four different pyramid partitions.
Last, we concatenate these context feature maps:
\begin{ceqn}
\begin{align}
    \mathbf{Z} = \rm{concate}(\mathbf{Z}^{1}, \mathbf{Z}^{2}, \mathbf{Z}^{3}, \mathbf{Z}^{6}).
\end{align}
\end{ceqn}

\begin{figure}
	\includegraphics[width=0.5\textwidth]{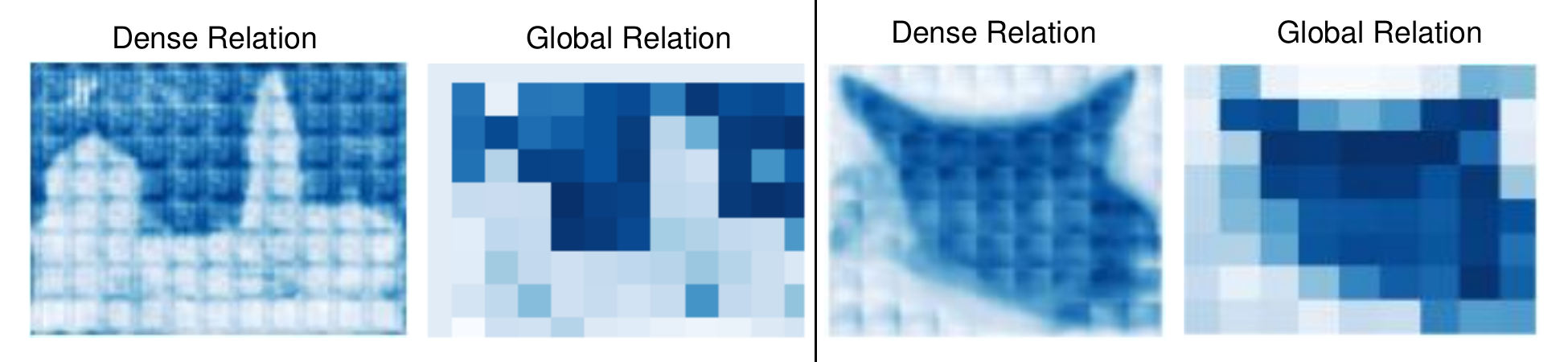}
	\caption{\small{
	Illustrating that the shape of the checkerboard in dense relation (based on interlaced sparse self-attention) is the same as the shape of the global relation map.
	The left / right two columns present the example from the $2$-rd / $1$-st row of Fig.~\ref{fig:dense_relation_1} (b) / (c) separately.
	}.
	}
	\label{fig:dense_relation_zoom}
\end{figure}

\vspace{2mm}
\subsection*{\textbf{I. Checkered artefact with ISA.}}
We can observe checkered artefact in the Fig.~\ref{fig:dense_relation_1} and Fig.~\ref{fig:dense_relation_2},
which is caused by our implementation on
the visualization of the global relation, local relation and dense relation.
We illustrate the related pseudo-code in Algorithm~\ref{fig:dense_from_global_local}.
Specifically speaking, for a selected pixel, we multiply a set of global relation matrices (associates with the pixels that belong to the same group as the selected pixel in the local relation stage) with the its local relation matrix.
Therefore, the shape of the checkerboard is exactly the same as 
the shape of the global relation map.
For example, 
we zoom in the dense relation and global relation of
some examples in the Fig.~\ref{fig:dense_relation_zoom}.

{
\lstset{
backgroundcolor=\color{white},
basicstyle=\fontsize{7.5pt}{8.5pt}\fontfamily{lmtt}\selectfont,
columns=fullflexible,
captionpos=b,
commentstyle=\fontsize{8pt}{9pt}\color{codegray},
keywordstyle=\fontsize{8pt}{9pt}\color{codegreen},
stringstyle=\fontsize{8pt}{9pt}\color{codeblue},
frame=none,
otherkeywords = {self},
autogobble=true,
breaklines=true,
numbers=left,
stepnumber=1,    
firstnumber=1,
numberfirstline=true
}
\begin{algorithm}
\tiny
\begin{lstlisting}[language=python]
def VisualizeRelation(i, j, H, W, P_h, P_w, Wg, Wl):
# i, j: indices of the pixel to be visualized
# H, W: the height and width of the input of ISA module
# P_h, P_w: Number of groups along H and W dimension
Q_h, Q_w = H // P_h, W // P_w
# Wg: global relation with shape [P_h*P_w, Q_h*Q_w, Q_h*Q_w]
# Wl: local relation with shape [Q_h*Q_w, P_h*P_w, P_h*P_w]

# obtain the indices of current pixel
global_h_idx, local_h_idx = i % P_h, i // P_h
global_w_idx, local_w_idx = j % P_w, j // P_w
global_idx = gobal_h_idx * P_w + global_w_idx
local_idx = local_h_idx * Q_w + local_w_idx

# global relation for current pixel
global_rel = Wg[global_idx, local_idx]          # [Q_h*Q_w]

# local relation for current pixel
local_rel = Wl[local_idx, global_idx]           # [P_h*P_w]

# dense relation for current pixel
multi_global_rel = Wg[:, local_idx]             # [P_h*P_w, Q_h*Q_w]
dense_rel = multi_global_rel * local_rel.reshape(P_h*P_w, 1)
dense_rel = dense_rel.reshape(P_h, P_w, Q_h, Q_w)
dense_rel = dense_rel.permute(0, 2, 1, 3).reshape(H, W)

return global_rel, local_rel, dense_rel
\end{lstlisting}
\caption{\small{Python code of visualizing the relation matrix.}}
\label{fig:dense_from_global_local}
\end{algorithm}
}

\end{document}